\documentclass{harvard-thesis}
\usepackage{rotating}
\usepackage{dblaccnt}
\usepackage{bibentry}
\nobibliography*
\setcitestyle{square}

\usepackage[final]{pdfpages}

\usepackage{tcolorbox}
\usepackage{xcolor}
\usepackage{xspace}
\usepackage{soul}

\usepackage{csquotes}

\graphicspath{{/Users/mai/Documents/Recherche/Bibliographie/MyWriting/PhdThesis2013/NguyenSMthesis/figures/},{/Users/mai/Documents/Recherche/Bibliographie/MyWriting/2021KiManouryDuminy/5ee2d057d4158d0001ebf599},{/Users/mai/Documents/Recherche/Bibliographie/MyWriting/2024IclrZadem/65093f37440f1c57bfea3f69},{/Users/mai/Documents/Recherche/Bibliographie/MyWriting/2021DuminyYumiAS/MdpiProofread}}

\let\OldMacro\minitoc
\renewcommand{\minitoc}{\vspace{-7em}{\OldMacro}\vspace{1em}}

\renewcommand*{\bibfont}{\normalfont\footnotesize}

\let\OLDthebibliography\thebibliography
\renewcommand\thebibliography[1]{
  \OLDthebibliography{#1}
  \setlength{\parskip}{0pt}
  \setlength{\itemsep}{2.7pt plus 0.2ex}
}

\usepackage{zref-user}
\makeatletter
\zref@newprop{chapter}{chapter \thechapter}
\zref@addprop{main}{chapter}
\zref@newprop{Chapter}{Chapter \thechapter}
\zref@addprop{main}{Chapter}
\zref@newprop{section}{section \thesection}
\zref@addprop{main}{section}
\makeatother

\newcommand{\real}[0]{\ensuremath{\mathbb{R}}}
\newcommand{\taskgoal}[0]{\ensuremath{g^{*}}}
\newcommand{\dist}[1]{\ensuremath{\lVert {#1} \rVert_2}}

\newcommand{\abs}[0]{\mathcal{N}}

\newcommand{\states}{\ensuremath{\mathcal{S}}}
\newcommand{\partitions}{\ensuremath{\mathcal{G}}}
\newcommand{\partition}{\ensuremath{G}}
\newcommand{\edges}{\ensuremath{\mathcal{E}}}

\newcommand{\policy}{\ensuremath{\pi}}
\newcommand{\navpolicy}{\ensuremath{\policy_{\text{Comm}}}}
\newcommand{\manpolicy}{\ensuremath{\policy_{\text{Tut}}}}
\newcommand{\contpolicy}{\ensuremath{\policy_{\text{Cont}}}}

\newcommand{\navigator}{\textit{Commander}\xspace}
\newcommand{\manager}{\textit{Tutor}\xspace}
\newcommand{\controller}{\textit{Controller}\xspace}

\renewcommand{\hl}[1]{#1}

\begin{document}

\title{The intrinsic motivation of reinforcement and imitation learning for sequential tasks}
\subtitleg{Body movement analysis, activity of daily living analysis and  intrinsically motivated robot for hierarchical reinforcement learning  }
\titlefr{La motivation intrinsèque d'apprendre par renforcement et imitation des tâches séquentielles}
\author{Nguyen Sao Mai}
\advisor{}

\degree{Habilitation à Diriger des Recherches}
\field{Engineering Science}
\degreeyear{2024}
\degreemonth{}

\department{Computer Science, Data and AI }
\university{Sorbonne Université }
\universitycity{Paris}
\universitystate{France}

\maketitle
\copyrightpage
\resumepage 
\dominitoc
\tableofcontents
 
\setcounter{page}{1}
	\pagenumbering{arabic}

\onehalfspacing
\clearpage

 \newpage
\chapter{Approach and State of the Art}

My research has been investigating in developmental cognitive robotics \citep{Cangelosi2015,Asada2009ITAMD,Lungarella2003CS}, the motivations of learning agents to interact with teachers, with the outlook of a fully autonomous robot that must continuously learn new tasks and how to perform more complex tasks relying on simpler ones, so as to adapt to the human daily environment and user's needs that change constantly. 
The paradigm of a permanent learning process is called \emph{lifelong learning} or \emph{continual learning} \citep{Ring1994}. 

More specifically,  learning in an unbounded environment a non-finite number of tasks  is referred to as  \textit{open-ended learning} \citep{Doncieux2018FN}.  Open-ended learning is characterised by multi-task learning, where the set of tasks is not known in advance, including interrelated tasks where easy tasks can be composed into more complex tasks. Thus state, action and task spaces are a priori unbounded in time, space and complexity, thus can be of infinite dimensionality. 
 The \textbf{challenges of open-ended learning are that (1) the state and action spaces are continuous and high-dimensional, even of infinite dimensionality, and  (2) the mappings to learn can be stochastic and redundant}. 

In this chapter, I will describe the developmental learning approach, the challenge and outlook taken by our works, and the relations of our works in active imitation learning with the state of the art.
 
\section{Developmental Robotics}

Uncovering the mechanism babies acquire new skills has been proposed as a key to enable artificial intelligence agents, since Alan Turing suggested :

\begin{displayquote}
Instead of trying to produce a program to simulate the adult mind, why not rather try to produce one which simulates the child's? If this were then subjected to an appropriate course of education, one would obtain the adult brain [...] Our hope is that there is so little mechanism in the child brain that something like it can be easily programmed. The amount of work in the education we can assume, as a first approximation, to be much the same as for the human child. 

(Turing, A.M. (1950). Computing machinery and intelligence. Mind, LIX(236):433–46).
\end{displayquote}
 
Taking Turing's suggestion, \emph{developmental robotics}, first introduced in \citep{Asada2001RAS}, proposes an “interdisciplinary approach to the autonomous design of behavioral and cognitive capabilities in artificial agents (robots) that takes direct inspiration from the developmental princi­ples and mechanisms observed in the natu­ral cognitive systems of ­ children”, as phrased in \citep{Cangelosi2015}. The main goal is to uncover the principles regulating the acquisition process of an increasingly complex set of sensorimotor and cognitive skills, through real-time interaction between its brain, its body and its environment \cite[Chapter~3]{Cangelosi2022}. This  online, open-ended, cumulative learning process is generally viewed as a progressive evolution of abilities and skills, akin to the developmental stages of infants as described by \cite{Piaget1952}, which principle derived into the machine learning fields of \emph{curriculum learning}  \citep{Bengio2009P2AICML}.

Part of the  highly interdisciplinary field of cognitive science which includes developmental psy­chol­ogy, neuroscience, robotics, linguistics and computer science,  developmental robotics is centered around two key concepts:
\begin{itemize}
\item \textbf{enactivism}, introduced by \citet{Varela1991},  hypothesises that cognition is based on situated, embodied agents. It is a  theoretical approach to understanding the mind which emphasises the way that organisms and human mind organise themselves by interacting with the environment. Enactivism thus uses the notion of \emph{embodiment} \citep{Brooks1991AI,Pfeifer1999} and action for cognition, which hypothesises that the mind is largely determined by the form of the organism's body and the actions it can do. The embodied cognition is grounded on self-experience. 
The way we conceptualise and reason depends on "the kinds of bodies we have, the kinds of environments we inhabit, and the symbolic systems we inherit, which are themselves grounded in our embodiment”\citep{Johnson1987}.  The mind builds from the personal history of each agent of his sensorimotor perception, and is not mere manipulation or operations of  abstractions. This translates Piaget’s theory emphasis on the sensorimotor bases of mental development and on the environmental approach balancing the biological approach.  
Drawing a parallel with machine learning, this learning paradigm is the closest to reinforcement learning, where the agent learns a policy by interacting with its environment to learn online from reward signals while discovering its environment.

\item the \textbf{social learning theory} proposes that new behaviors can be acquired by direct instructions or observing and imitating others \citep{Bandura1971}.  The learning benefits from the observation of behavior, but also the observation of rewards and punishments. This learning paradigm relies on data induced by a teacher. Vygotsky's sociocultural theory \citep{Vygotsky1978} describes the social environment's role as  important  as the physical environment  on the scaffolding of the child’s cognitive system owing to the role of adults and peers in guiding the development of the child. Educational psychology has outlined the importance of social learning to bootstrap autonomous learning, and put forward the idea of \emph{social scaffolding} : children explore and learn on their own, but in the presence of a teacher, they can take advantage of the observations of  social cues and the teacher's behavior to accomplish more.  A teacher often guides a learner by providing timely feedback or instructions, luring them to perform desired behaviours, and controlling the environment to simplify the analysis of the state, for instance by highlighting the appropriate cues to make them salient. \citet{Vygotsky1978} defines the concept of \emph{zone of proximal development} as the space between what a learner is capable of doing autonomously and what the learner cannot do even with support : 

\begin{displayquote}
$[$The zone of proximal development$]$ is the distance between the actual developmental level as determined by independent probem solving and the level of potential development as determined through problem solving under adult guidance or in collaboration with more capable peers.

\citet{Vygotsky1934}
\end{displayquote}

 Backed by findings on mirror neurons in the brain \citep{Rizzolatti2008} that suggests a natural predisposition for imitation, computational models of imitation have been developped such as in \citep{Oztop2006NN}. More broadly, social learning has contributed to  human-robot imitation studies such as imitation, gesture, vocalization,  joint attention, turn taking, etc. In machine learning, this paradigm translates in its simplest form as supervised learning, with a wealth of algorithms named \emph{imitation learning} \citep{Nehaniv2007,Schaal1999TCS}, programming by demonstration \citep{Calinon2009,Billard2007RobotProgrammingby} or learning from demonstration \citep{Argall2009RAS}.
\end{itemize}

Our works falls within the developmental robotics approach, to propose machine learning algorithms, with a focus on motor control and action learning. 

\section{Reinforcement Learning and Imitation Learning}

From the machine learning perspective too, these two key concepts seem to shape two families of  action learning algorithms. They can be seen as pertaining to the trial-and-error paradigm, where the learning agent collects data by exploring its physical environment autonomously and observing the outcomes of its actions, or to the imitation learning, where the learning relies on data provided by a tutor.

While imitation learning has been investigated to learn a policy for an Markov Decision Process given expert demonstrations, for a long time in robotics \citep{Schaal2003PTRSLSBS,Billard2004RAS},  socially guided  learning has been strongly relying on the involvement of the human user or the demonstration dataset. However, the more dependent on the human, the more challenging learning from interactions with a human is, due to limitations of the provided demonstration dataset because of lack of human patience, attention, memory, or the sparsity of teaching datasets, the absence of teaching for some subspaces, ambiguous and suboptimal human input, correspondence problems, etc, as highlighted in \citep{Nehaniv2007}. This is one of the reasons why in most approaches to robot learning of motor skills, either in supervised learning or inverse reinforcement learning, only a few movements or motor policies were learnt in any single studies. More recently, behavioural cloning has been implemented using deep neural networks, as reviewed in  \citep{Mandlekar2021APA}, but show the same limitations and a reliance on a higher amount of demonstration data. In particular, imitation learning has been performing poorly for multi-task learning, including hierarchical tasks.  A limitation is that as the number of tasks increases, the number of data needed  and the required quality of data from tutors increase, thus  increasing the cost of human tutoring. Robot learning algorithms need to decrease the cost of human guidance, by both leveraging demonstrations on simple tasks to learn more complex tasks, and by reducing the cost for human tutors of  the interactions.
 \textbf{Increasing the learner's autonomy from provided demonstrations could address these limitations}.

On the other side, the trial-and-error strategy has been implemented in computational models as reinforcement learning algorithms \citep{Sutton1998} to infer the optimal behaviour, which is also referred to as the \emph{policy}. To improve the efficiency of exploration, methods reusing some of the concepts elaborated in the statistical active learning framework \citep{Fedorov1972,Cohn1996JAI,Roy2001P1ICML}  have recently been developed in the fields of developmental  learning using a heuristics  for its active choice \citep{Oudeyer2009FN,Schmidhuber2010ITAMD,Gottlieb2013TCSTCS}. This heuristics is inspired by \textit{intrinsic motivation} in psychology. 
Intrinsic motivation was described in \citep{White1959PR} : 
``While the purpose is not known to animal or child, an intrinsic need to deal with the environment seems to exist and satisfaction (the feeling of efficacy) is derived from it.''
Intrinsic motivations  triggers spontaneous exploration and curiosity in humans. 
\begin{displayquote}
 ``Intrinsic motivation is defined as the doing of an activity for its inherent satisfaction rather than for some separable consequence. When intrinsically motivated, a person is moved to act for the fun or challenge entailed rather than because of external products, pressures or reward''
 \citep{Ryan2000CEP}
 \end{displayquote}

 Intrinsic motivation is to be contrasted with extrinsic motivation, which is ``a construct that pertains whenever an activity is done in order to attain some separable outcome. Extrinsic motivation thus contrasts with intrinsic motivation, which refers to doing an activity simply for the enjoyment of the activity itself, rather than its instrumental value''~\citep{Ryan2000CEP}.

 Still, the current reinforcement learning algorithms fail to transpose to robots because: (1) they require the robot to execute a large number of real actions to sample the continuous high-dimensional space of states and actions, hence requiring a lot of time and are not scalable for robots; (2) they cannot leverage the learned tasks to compose more complex tasks for open-ended learning, as the more sequential the considered tasks, the higher dimensionality the state-action spaces are. Therefore,  \textbf{complementary developmental mechanisms need to constrain the growth of the size and complexity of the exploration areas and structure the environment and the learning curriculum}, by guiding them toward learnable subspaces and away from unlearnable subspaces. 
 
While reinforcement learning and socially guided learning have so far often been studied separately in developmental robotics and robot learning literature, we believe their integration has high potential. Their combination could push the respective limits of each family of exploration mechanisms.  I argue that \textbf{social guidance, leveraging knowledge and skills of others, can be key for bootstrapping the intrinsically motivated learning of such models}.

\section{Framework}

Let us outline a framework for learning open-ended problems from reinforcement learning and imitation learning.

\subsection{Formalising the learning problem}

\label{sec:formalisationIntro}
Let us consider a robot interacting with a non-rewarding environment by performing sequences of motions of unbounded length in order to induce changes in its surroundings.

 \begin{figure}[tb]
  \centering
  \includegraphics[width=1\textwidth]{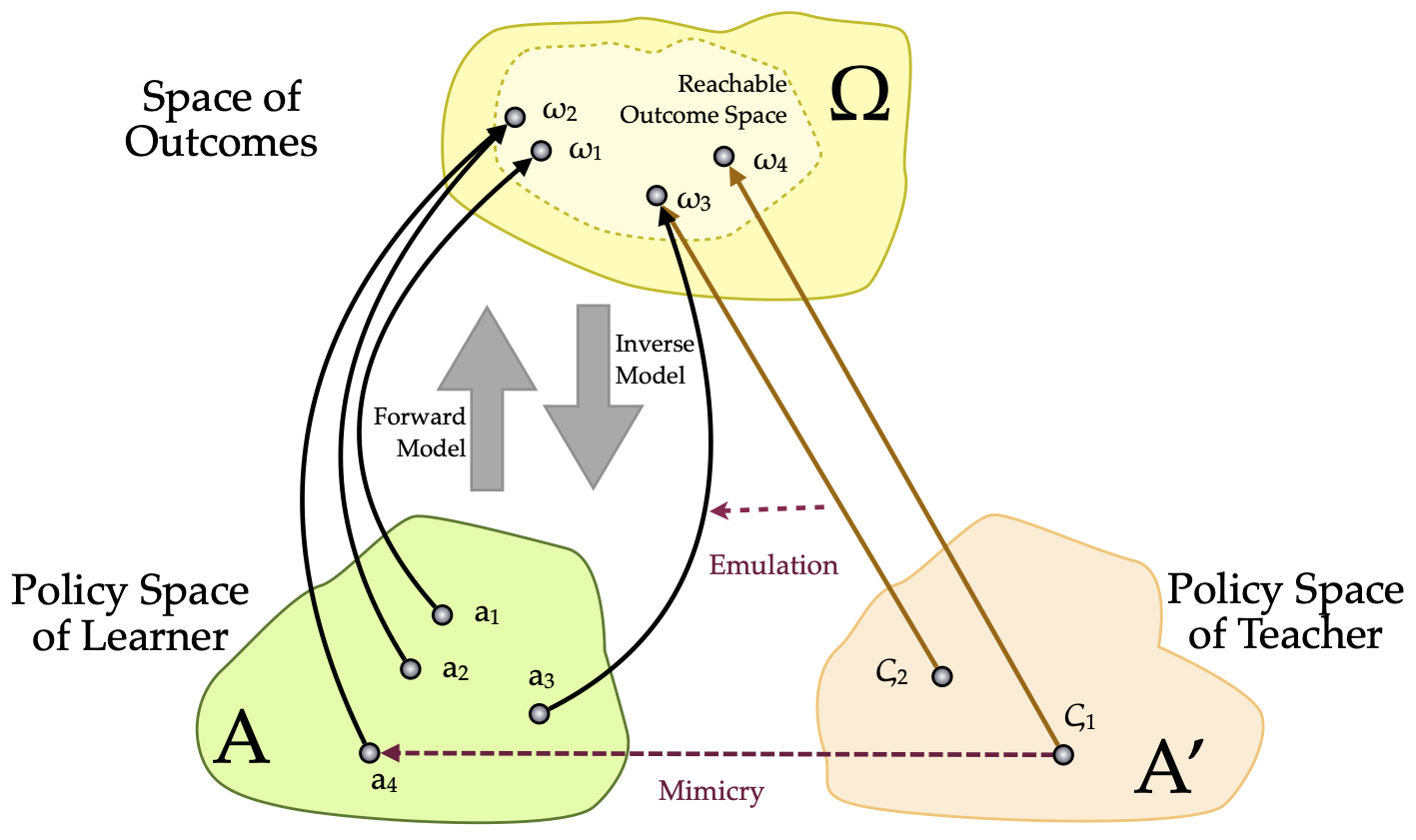}
\caption{The goal is to learn a mapping from the policy space $A$ of the learner and the outcome space $\Omega$. The mapping from the policy space to the outcome space is the forward model, used for predicting the effects of an action. The mapping from the outcome space to the policy space is the inverse model, used to choose the appropriate action to reach the goal. Emulation and mimicry are two types of imitation learning. In emulation, the learner tries to reproduce the outcome $\omega_3$ demonstrated by the teacher without trying to reproduce the teacher's movement $\zeta_2$, but uses its own movement $a_3$. In mimicry, the learner reproduces the demonstrated movement $\zeta_1$ with policy parameters $a_4$, without trying to reproduce the demonstrated outcome $\omega_4$.}
\label{fig:formalisationSGIMACTS}
\end{figure}

Each of these motions is named a \textit{primitive action}, described by a parametrised function with $p$ parameters: $a \in \mathcal{A} \subset \mathbb{R}^p$. We call the continuous multidimensional space $\mathcal{A}$ the \textit{primitive action space}. 
Our robot can perform sequences of primitive actions.

The actions performed by the robot have consequences on its environment, which we call outcomes $\omega \in \Omega$, where $\Omega$ is a subspace of the state space $S$ defining the control tasks to learn. Once the robot knows how to cause an outcome $\omega$, we say the outcome is then \textit{controllable}. The set of controllable outcomes is  $\Omega_{cont} \subset  \Omega$ and this set changes as the robot learns new tasks. For convenience, we define the \textit{controllable space}  $\mathcal{C} = \mathcal{A} \cup \Omega_{cont}$, regrouping both primitive actions $\mathcal{A}$ and observables that may be controlled, $\Omega_{cont}$.

The robot learns tasks/models $T$, each mapping a controllable $c \in \mathcal{C}_T \subset \mathcal{C}$ to an outcome $\omega \in \Omega_T \subset \Omega$ within a given context $s \in S_T \subset S$. 
More formally, a task is a set of: a \textbf{forward model} $M_T:(S_T,\mathcal{C}_T) \to \Omega_T$ and an \textbf{inverse model} $L_T:(S_T,\Omega_T) \to \mathcal{C}_T $.  The forward model is used to predict the observable consequence ${\omega}$  from a given context $s$ of a controllable $c$, which is either a primitive action or a goal state that can be induced by a compound action. Conversely, the inverse model is used to estimate a controllable $\tilde{c}$ to be performed in a given context $s$ to induce a goal observable state $\tilde{\omega}$: if $\tilde{c}$ is a primitive action the robot executes the action, otherwise it sets $\tilde{c}$ as a goal state and infers the necessary actions using other inverse models. 

These models are trained on the data recorded by the robot along its exploration in its dataset $\mathcal{D}$. We define a strategy $\sigma$ any process for exploration. For instance, we consider autonomous exploration and imitation learning strategies. Formally, we define it as a data collection heuristics, based on the current data, that outputs a set of triplets of initial state, action and outcome: $\sigma : \mathcal{D} \mapsto\{ (s, a, \omega) \}$.

\subsection{Formalising the imitation process}

Imitation learning can be separated into two broad categories of social learning  \citep{Call2002Threesourcesof}, as represented in fig. \ref{fig:formalisationSGIMACTS} :  the mimicry strategy where the learner copies the observed trajectory $\zeta_1$ with policy parameters $a_4$, and emulation strategy where the learner attempts to replicate goals $\omega \in \Omega$. 

Several teachers generate trajectories according to their policies $\pi_k$, which are a priori unknown to the learner. A priori, the action space of the teacher  is not equal to the action space of the learner. We suppose that  $\pi_k$ depends on the current state, the goal, but can also be influenced by the query of the learner. In the next sections, we discuss the query process from a learner to a tutor.

\section{Active Imitation Learning}

While reinforcement learning and imitation learning have been traditionally opposed -- the former being typically a data collection algorithm through exploration of the environment, and the latter typically using a preset dataset of labelled data -- we show that both worlds are on the contrary complementary and highlight the merits of merging the two fields. While an increasing stream of machine learning works propose to combine the two paradigms \citep{Price1999I,Ng2000P1ICML}, and has been named \emph{interactive reinforcement learning} \cite{Thomaz2005A2WHCML,Arzate-Cruz2020P2DISC}. This stream of work includes reinforcement learning from human feedback \citep{Thomaz2008AIJ,Knox2013LNCS,Christiano2017ANIPS} including for large language models \citep{Ouyang2022ANIPS}. In most interactive reinforcement learning works, the agent in these works undergoes the interaction passively. 
We show how a learning agent could optimise its interaction with tutors. 

Research in developmental psychology indicates that social interaction is  not only for social pleasure but can serve for learning and exploration of the environment. While most theories of infant social learning focus on how infants learn whatever and whenever the adults decide to teach them, recent findings suggest that social learning is not a passive process but that infants play an active role in collecting information and adapting their learning strategy according to their interests. Indeed,   infants show a preference to learn from reliable people \citep{Poulin-Dubois2011IBD}, and their attention towards adults is influenced by the adult's role in bringing new information \citep{Begus2016PNAS}. Infants show curiosity and active contribution to social transmission of knowledge \citep{Begus2018ALFICSMCLMCuriousLearners:How}. Thus, we propose to frame  the interaction with human teachers as a reinforcement learning problem. Reinforcement learning is therefore not only about an agent interacting with a physical environment but also with a social environment. Could the reinforcement framework be extended to all aspects of the interaction with teachers?

In the next section, we review the literature describing on how a learning agent  devises its learning strategy between the two learning paradigms and aspects of its interaction with its tutors.   We will refer the approaches combining autonomous exploration and social guidance in an active way as \emph{active imitation learning}, a term defined in \citep{Shon2007AAAI}, as an  ' imitation learning paradigm covering cases where an observer can influence the frequency and the value of demonstrations that it is shown'. This corresponds to a more general definition than the \emph{ask for help} approach where 'The learner is completely responsible for its interaction with the trainer, asking the trainer to provide actions. The trainer will always give an action when asked' \citep{Clouse1996}. Later on, \cite{Judah2014JMLR} defined used active imitation learning, in a more restricted case where 'the learner asks queries about specific states, which the expert labels with the correct actions. The goal is to learn a policy that is nearly as good as the expert’s policy using as few queries as possible'.

\subsection{Optimising What, How, When and Who to imitate}
\label{sec:SocialGuidanceReview}

In order to define the social interaction that we wish to consider, let us characterise the different possibilities of information flow  as reviewed in \citep{Argall2009RAS,Billard2007RobotProgrammingby,Schaal2003PTRSLSBS,Lopes2009AbstractionLevelsfor} with respect to: what, how, when and who to imitate. This categorisation has been introduced in \citep{Dautenhahn2002, Breazeal2002TCS}. 
In this section, we examine the related works studying the closed-loop phenomenon of a learner making queries to a tutor that influence what data are added to its training set. 
Compared to the review by \cite{Zhang2019PTIJCAII} which focuses on the deep reinforcement learning and analyses the state of the art with respect to the types of learning algorithm, we propose  to analyse in this section the existing works under the human-robot interaction perspective. As opposed to the more general review in  \citep{Ravichandar2020ARCRAS}, we will focus on active choices by the learner on its interaction with teachers, and contrast our works to the state of the art.

  Please note that the current review does not aim at friendly human-robot interaction, the social rules for a comfortable and natural interaction or goal understanding and intentionality. 
  
In this section,  we focus on the information from the tutor to the learning agent and its  efficiency to convey content by human-robot interaction. Thus we examine what information is given by the tutor.

\begin{sidewaystable}
\small
\begin{tabular}{|p{2.2cm} |p{2.1cm} |p{1.85cm} |p{1.5cm}  |p{1.5cm}  |p{1.5cm}  |p{3.5cm}  |p{2.8cm}  |p{2.cm}  |}
\hline
Algorithm & Environ. & \multicolumn{4}{|c|}{Imitation} & \multicolumn{3}{|c|}{Query} \\ 
~ & Tasks & What & How & When & Who & What  & When & Who \\
\hline
AfH \citep{Clouse1996} & Single &
  Policy & Low-level  & Learner's initiative & Unhelpful & Difference highest-lowest Q values & NA & NA \\
\hline
\citep{Shon2007AAAI} & Single & Discrete States & Low-level &Learner's Initiative & Unhelpful, several & NA & Value of demonstration & Policy regret\\
\hline
CBA \citep{Chernova2009JAIR} &Single & Discrete Policy & Corrective demonstration & Learner's initiative & Helpful, one & Current state &   Confidence on state & NA \\
\hline
Dagger \citep{Ross2010}
& Single & Policy & Low-level  & After each episode & Helpful & Action for the visited states & NA  &  NA\\
\hline
FPS \citep{Argall2011RAS} & Continuous Tasks of different complexities & Trajectory segment & Advice, Reward & Always & One, helpful & Feedback on visited trajectories &  NA & NA  \\
\hline
CLIC \citep{Fournier2019ITCDS} & Multiple discrete & discrete Task, policy & Low-level & Fixed frequency & Unhelpful &  Task with highest learning progress & NA & NA\\
\hline
 \citep{Kulak2023ITCDS}& Single continuous & Continuous policy & NA & Learner's initiative & Helpful, one & Goal based on epistemic uncertainty &  expected epistemic entropy reduction  & NA \\
\hline

RPI \citep{Liu2024TICLR} & Single & Continuous policy & Low-Level & Learner's initiative & Several & Action for random state in trajectory &    Highest Value & Highest Value \\
 \hline
SGIM-ACTS \citep{Nguyen2012PJBR} 
&Multiple continuous & Continuous policy, Continuous outcome & Low-level  & Learner's initiative & Unhelpful, multiple & Action for the outcome with highest competence progress & Competence progress & Tutor enabling highest competence progress \\ 
\hline
SGIM-PB \citep{Duminy2021AS} 
& Continuous tasks of different complexities & Policy, Outcome, task decomp & Low-level  & Learner's initiative & helpful, multiple & Action for the outcome with highest competence progress & Competence progress & Tutor enabling highest competence progress \\
\hline
\end{tabular}
\caption{Comparison of the active imitation learning algorithms: aspects of the imitation and of the interaction the learner actively decides on.}
\label{tab:activeImitation}
\end{sidewaystable}

\begin{itemize}

\item {What to imitate and query?}
Let us  examine the target of the information given by the teacher, or mathematically speaking, the space on which he operates. This can be either the policy, context or outcome spaces, or combinations of them. 
  Merging autonomous exploration with socially guided exploration can optimise the number of data needed for multi-task learning by reusing knowledge across tasks by interpolating for parameterised tasks or using hindsight replay \citep{Nguyen2012PJBR}, and by querying and combining different types of demonstrations such as policy demonstration, goal demonstration and task decomposition \citep{Nguyen2014AR,Duminy2019FN}.  The learner can query demonstrations for all states visited in the previous episode \citep{Ross2010}, or for a state with a low difference between the highest and the lowest Q values \cite{Clouse1996}. In \citep{Fournier2019ITCDS} and in our works \citep{Nguyen2011IICDL,Duminy2018ICSC}, the learner queries a demonstration for the task with highest competence progress. 
  
\item How to imitate ? 
  Whichever the target, the information can be communicated from the teacher to the learner through low level (as in our algorithm SGIM-ACTS \citep{Nguyen2014AR}) or high level demonstrations, through advice, reward or labelling.  A language protocol often enables instructions of policies  \citep{Argall2011RAS}. Language become a cognitive tool to support efficient acquisition of open-ended repertoires of skills as described in \citep{Colas2019P3ICML}. 

\item When to imitate ?
 The timing of the interaction varies with respect to its timing within an episode, and with respect to its general activity during the whole learning process. Demonstration data can be used at the beginning of the learning process to initialise the model, at a fixed or a random frequency, at the teacher's initiative or at the learner's initiative. When at the learner's initiative, it optimises the interaction with human supervisors, so as to minimise the cost of human guidance in terms of number of demonstrations. The learner query teachers' guidance based on criteria such as uncertainty \cite{Chernova2009JAIR} and in particular epistemic uncertainty  \cite{Kulak2023ITCDS}. Our works used a criteria competence progress to select the best strategy between autonomous exploration and the imitation strategies \citep{Nguyen2012PJBR, Nguyen2012IICHR, Duminy2019FN}, 

\item Who:  Being able to request help to different experts is also an efficient way to address the problem of the reliability of the teacher. Imitation learning studies often rely on the quality of the demonstrations, whereas in reality a teacher can be performant for some outcomes but not for others. Demonstrations can be ambiguous, unsuccessful or suboptimal in certain areas. Like students who learn from different teachers who are experts in the different topics of a curriculum, a robot learner should be able to determine its best teacher for the different outcomes it wants to achieve.  To our knowledge, \cite{Shon2007AAAI} were the first to propose a framework to enable the learning agent to decide who to imitate, using a side payment. \cite{Liu2024TICLR} proposed a criteria by comparing the value functions of each teacher's policy. Our works \citep{Nguyen2012PJBR,Duminy2019FN,Duminy2021AS} make the most of several teachers by choosing the expert for each query based on the competence progress each teacher has enabled.
Learners also minimise the cost of human guidance by considering their availability through a cost parameter to choose whom to ask guidance to. 
These methods are robust to imperfect teachers \cite{Shon2007AAAI,Fournier2019ITCDS,Nguyen2013IICDLE,Nguyen20122IISRHIC,Duminy2016I2JIICDLER}. 

\end{itemize}

A synthesis of how the state-of-the-art tackles these aspects of active imitation learning is presented in Table \ref{tab:activeImitation}.

While learning single tasks can be addressed by naive methods, continuous task spaces require more efficient reinforcement learning algorithms or models. 
Further, state-of-the-art methods are the most profitable for multi-task learning where tasks of different nature constitute composite continuous task spaces. 
 In particular, hierarchical/sequential task learning can combine transfer knowledge from simple tasks to complex tasks after identifying through learning the relationships between tasks \citep{Duminy2021AS}, thus entailing automatic curriculum learning that self-schedules the learning process with simple tasks before the agent starts to master more complex/sequential tasks.

\subsection{The Reward Hypothesis of Social Interaction}

Yet, no algorithms have yet proposed a comprehensive framework to optimise all types of social interaction. Most works have focused on addressing what and how to interact. Only a few have addressed when and who to imitate. A few works such as \citep{Nicolescu2003PSIJCAAMS,Thomaz2006} have combined several types of messages such as policy demonstrations, highlight on the context elements, corrective and reward feedback, but they rely on the teacher to optimise the interaction. The learning agent does not actively choose the aspects of this interaction.  

In active imitation learning, intrinsic motivation based algorithms such as CLIC \citep{Fournier2019ITCDS} or SGIM \citep{Nguyen2021KI} have proposed a common criteria to optimize what, how, when and who to imitate: the learning progress.  These optimisations rely on empirical estimations of the impact of additional data on the learning progress for each source of information, and on the active selection of the information to be queried. The algorithms also optimise the number of data needed for multi-task learning by reusing knowledge across tasks and by querying and combining different types of information, after identifying through learning the relationships between tasks.  The learning progress measure can be used to merge autonomous exploration with socially guided exploration, select the best among several teachers, and be robust to imperfect teachers. In particular, choosing aspects of the interaction with the teachers, such as who to imitate and when to imitate can be optimised with, as criteria, the learning progress a teacher can bring to the learner, weighted by the cost of querying a demonstration to the teacher. This criteria can thus be viewed as a reward to interact with teachers. 

While reinforcement learning and social interaction seem to live in different frameworks, our review shows that they can be combined. However a common framework is still needed.
In reinforcement learning,  
 \cite{Sutton2018} posited the reward hypothesis as :  
\begin{displayquote}
 “all of
what we mean by goals and purposes can be well thought
of as maximization of the expected value of the cumulative
sum of a received scalar signal (called reward).” 
 \citep{Sutton2018}
 \end{displayquote}

However, this reward hypothesis has only been tested in autonomous learning settings, and has not  been extended to settings where the learning agent interacts with teachers, who may be perfect or imperfect. How can a learning agent learn to choose right interaction actions with teachers ? \textbf{Which reward function should shape the learning agent's interactive behaviour ?}

While Human-Computer Interaction and Affective computing research postulate that the drive for interaction with others comes from emotions, thus emotional signals can act as an external reward signals, we take \textbf{the outlook of interaction as an active learning process to gain information, and the others as sources of information}, which probability distribution the learner has to model through sampling. Thus, we strive to \textbf{model the intrinsic motivation of interacting with teachers.}

 Recently, some works have proposed a universal reward function for reinforcement learning of social interaction : the learning progress. In \citep{Fournier2019ITCDS,Nguyen2021KI},  {the same formulation of the intrinsic motivation  is valid both for autonomous exploration and  for social guidance, may the demonstrations requested to tutors be low-level policies, goals or decomposition into subgoals}. Thus, a common framework is possible by \textbf{using intrinsic motivation as the common criteria to choose each learning strategy}. Our works shows that by considering that \textbf{each choice of teacher and type of demonstration is a learning strategy, the common criteria to choose a learning strategy $\sigma$ and task $\omega$ can be formulated as  $im(\sigma, \omega) = \kappa(\sigma) * progress(\sigma, \omega) $} where $progress$ is the empirical progress measured through the last episodes with strategy $\sigma$ and goal $\omega$, and $\kappa(\sigma)$ is the cost of the strategy, representing the availability of teachers, their willingness to interact with the robot ... 

At the convergence of active imitation learning and open-ended learning in non-rewarding environments, we describe in the next chapter our multi-task learning algorithms  for multiple parametrised tasks and for hierarchical tasks, using intrinsic motivation. In chapter 3, we detail our active imitation learning algorithms. In chapter 4, we describe  application projects in socially assistive robotics and assistive technologies. Lastly, chapter 5 sketches future research directions.

\chapter{Intrinsic Motivation to Learn  Compositional Tasks}

In this chapter, we describe, for open-ended learning in environments without external rewards, our works to learn with active imitation learning without external rewards from the environment or the teachers, and to solve long-horizon tasks such as compositional tasks, also referred to as sequential tasks. 
While most works on multi-task learning focus on parametrised tasks or tasks of the same level of complexity, my team examined how
to learn \emph{compositional tasks}, ie tasks can be considered as composed of simpler tasks.

Indeed, humans, and other animals, cannot only learn simple behaviors and generalise them across various states, but they can also combine simple behaviors to create more complex behaviors.  This idea of decomposition and composition of  actions into sequences of reusable
primitives has been formalised  as the motor schemata theory by \cite{Arbib1981Perceptualstructuresand}. The
theory considers that a set of action primitives might be memorised, to be retrieved  and combined by the higher level  to generate desired actions. 
The
ability to compositionally combine behaviors is thought to be central to generalized
intelligence in humans and a necessary component for artificial intelligent systems. 
The ability to learn a variety of compositional, long-horizon skills while generalizing to novel concepts remains an open challenge. Long-horizon tasks demand sophisticated exploration strategies and structured reasoning, while generalization requires suitable representations.

Applying \emph{Reinforcement Learning} (RL)~\citep{Sutton1998} to the complex environments commonly found in robotics is often challenging, due to the high-dimensional and continuous state space and sparse rewards for temporally extended tasks, especially for \textit{open-ended learning} \citep{Doncieux2018FN}.

To solve open-ended learning, several successive representational redescription processes through unsupervised acquisition of a hierarchy of adapted representations is suggested in \citep{Doncieux2018FN} to be a key component.  Open-ended learning  should be solved by an unsupervised acquisition of a hierarchy of adapted representations. To tackle learning multiple tasks of complexities unknown a-priori, the robot needs to exploit the knowledge from simple tasks to compose tasks of growing difficulty. A hierarchical description of actions has been proposed  in neuroscience (eg \citep{Grafton2007HMS}) and in behavioural psychology (eg. \citep{Eckstein2021CCCSS}). This idea has been transcribed in machine learning : Hierarchical Reinforcement Learning (HRL)  \citep{Barto2003DEDS} is a recent approach for learning to solve long and complex tasks by decomposing them into simpler subtasks. HRL could be regarded as an extension of the standard Reinforcement Learning (RL) setting as it features high-level agents selecting subtasks to perform and low-level agents learning actions or policies to achieve them. Hierarchical Reinforcement Learning  tackles the environment's complexity by introducing a hierarchical structure of agents that work at different levels of temporal and behavioral abstractions : lower-level agents learn to solve easier tasks and higher-level agents learn to solve more complex tasks, using the low-level agents' policies. However, the reinforcement learning algorithms still need millions of learning samples to learn and are not scalable for robot learning.

In this section, we hypothesize that one critical element for solving such problems is the notion of compositionality. Hierarchical RL (HRL) has designed this compositionality as an ability to learn concepts and sub-skills that can be composed to solve longer tasks. However, acquiring effective yet general abstractions for hierarchical RL is remarkably challenging.   In this chapter, I only examine autonomous exploration paradigms, and not the imitation paradigm.

\section{Multi-Arm Bandit with Intrinsic Motivation}

 My team examines how an agent can learn both the policies for simple and sequential tasks and \textbf{the relationship between tasks for the purpose of transferring knowledge from  simpler tasks into  sequential tasks} and proposes an architectural architecture, Socially Guided Intrinsic Motivation for Sequence of Actions through Hierarchical Tasks (SGIM-SAHT), common to the implementations of algorithms IM-PB, CHIME, SGIM-PB. Table \ref{tab:DiffAlgo} outlines the differences between them.

\begin{table}[tb]
\begin{tabular}{|p{2cm} |p{7cm} |p{4cm} |p{2.5cm} |}
\hline
Algorithm & Strategies $\sigma$ & Tasks set & Composite Actions \\
\hline
IM-PB & Action exploration, Outcome space exploration, Task decomposition exploration & Static set & Subgoals sequence \\
CHIME & Action exploration, outcome space exploration & Dynamic set (emerging tasks) & Planning \\
SGIM-PB & Action exploration, outcome space exploration, Task decomposition exploration; active imitation & Static set & Subgoals sequence\\
\hline
\end{tabular}
\vspace{-0.2cm}
\caption{Differences between the 3 implementations of SGIM-SAHT to learn compositional tasks}
\label{tab:DiffAlgo}
\vspace{-0.5cm}
\end{table}

\begin{figure}
\centering
\includegraphics[width= \textwidth]{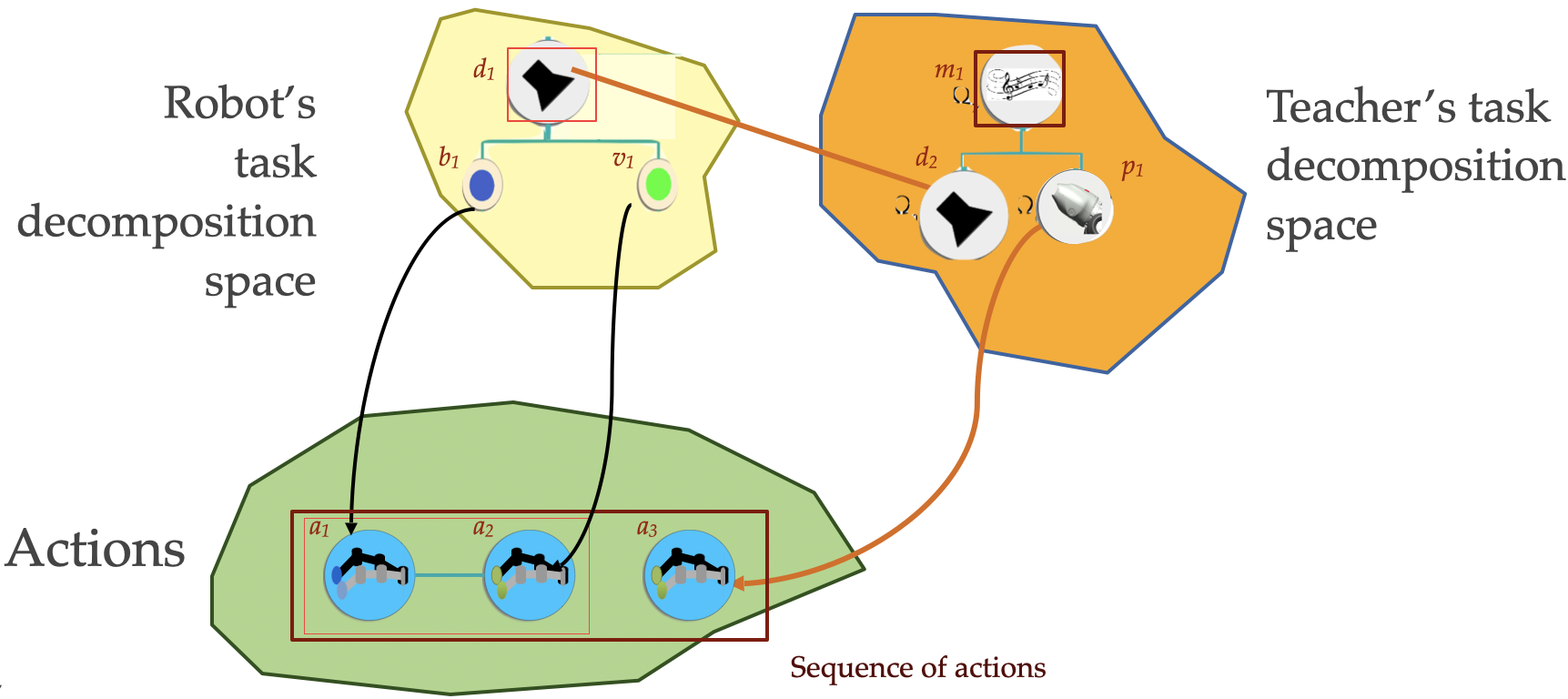}
\caption{ Our goal-oriented representation of sequential tasks decomposes a task into goals of subtasks. Each subtask can then recursively be decomposed into subtasks, to be in fine executed by a policy : $m_1$ is decomposed into $(d_2, p_1)$. $d_2$ is the decomposed into $(b_2,v_2)$  Thus a sequential task is in fine realised by a sequence of policies : $m_1$ is realised by the sequence $(a_1,a_2, a_3)$. When learning with a teacher, the learner can ask for a demonstration of low-level policy ($a_3$ realises $p_1$) or task decomposition ( $m_1$ is decomposed into subtasks $(d_2, p_1)$).
}
\label{fig:SequentialTasks}
\end{figure}

Our study relies on representations of task decomposition as illustrated in fig. \ref{fig:SequentialTasks}.  

\subsection{Formalisation}
\label{sec:formalisation}

We extend the formalisation proposed in Section \ref{sec:formalisationIntro} for sequential tasks. 

Our robot can perform sequences of primitive actions. Let a \textit{compound action} be a sequence of any length $n \in \mathbb{N}$ primitive actions, and be described by $n*p$ parameters : $a = [a_1, \dots, a_n] \in \mathcal{A}^n$. Thus the action space exploitable by the robot is a continuous space of infinite dimensionality $\mathcal{A}^{\mathbb{N}} \subset \mathbb{R}^{\mathbb{N}}$.

Let us note $\mathcal{H}$ the hierarchy of the models used by our robot. $\mathcal{H}$ is formally defined as a directed graph where each node is a task $T$ and its successors are the components of $ \mathcal{C}_T $.  As our robot learns this hierarchy, $\mathcal{H}$ varies along time.

\subsection{Algorithmic Architecture}

\begin{algorithm}[tbh]
    \caption{SGIM-SAHT \label{algorithm}}
    \begin{algorithmic}[1]
        \REQUIRE the different strategies $\sigma_1,...,\sigma_n$
        \REQUIRE the initial model hierarchy $\mathcal{H}$
        \ENSURE partition of outcome spaces $R \gets \bigsqcup_i \lbrace\Omega_i\rbrace$
        \ENSURE episodic memory \textit{Memory} $\gets \emptyset$
        \LOOP
        \STATE $\sigma, T, \omega_g  \gets$ Select Strategy, Task \& Goal Outcome($R, \mathcal{H}$) \label{algo:strategy}
        \STATE $l_c \gets$ Apply Strategy($\sigma, \omega_g$) \label{algo:sequence}
        \STATE $\mathcal{D} \gets  (\omega_r, a, l_c)  \gets$ Execute Sequence($l_c$) \label{algo:memory}
        \STATE  $(comp(\omega_g), comp(\omega_r)) \gets $ Compute Competence$(\omega_g,\omega_r))$\label{algo:competence}
        \STATE Update $M_T, L_T, \mathcal{H}$ with $(\mathcal{D}, comp(\omega_g), comp(\omega_r))$ \label{algo:M}
        \STATE $R_i \gets$ Update Outcome and Strategy Interest Map($R,\mathcal{D},\omega_g$)
        \ENDLOOP
    \end{algorithmic}
\end{algorithm}

\begin{figure}[t!]
\centering
\includegraphics[width=0.6\hsize]{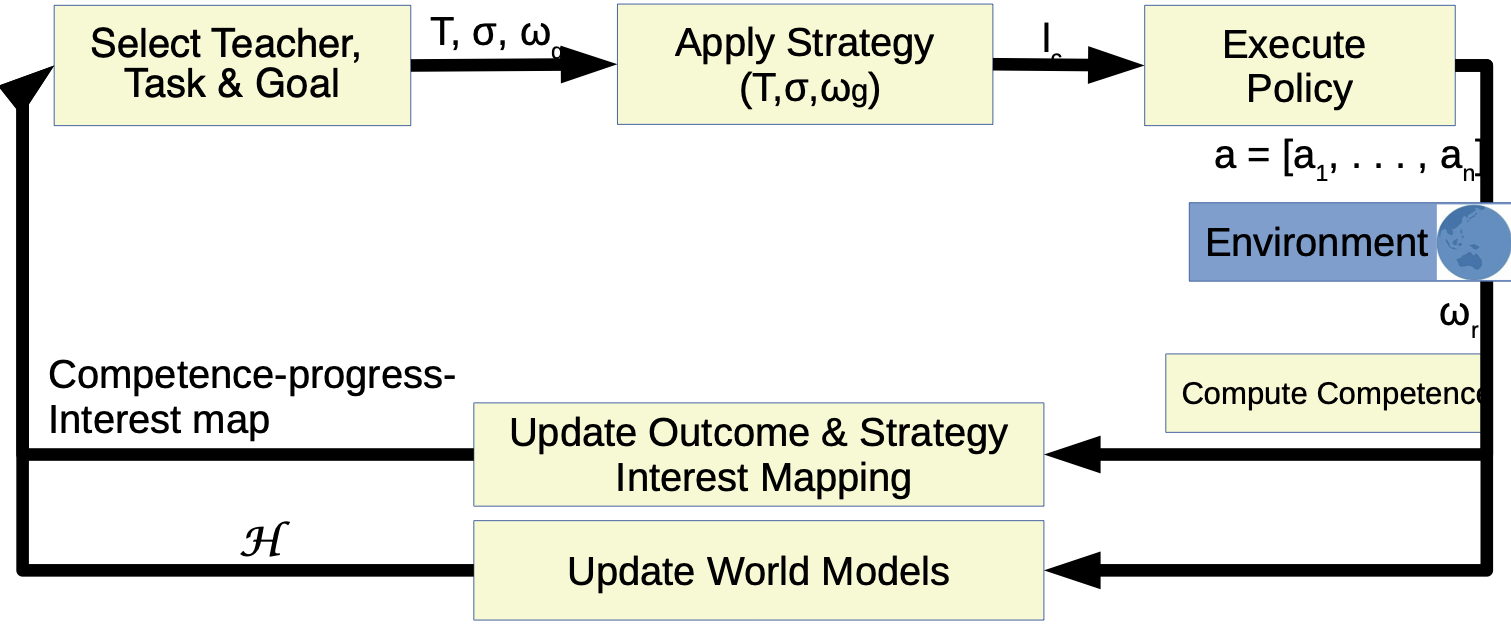}
\caption{The algorithmic architecture of SGIM-SAHT }
\vspace{-0.4cm}
\label{fig:algo}
\end{figure}

Socially Guided Intrinsic Motivation for Sequence of Actions through Hierarchical Tasks (SGIM-SAHT), detailed in Algo. \ref{algorithm} and outlined in Fig. \ref{fig:algo}, learns by episodes in which a task $T$ to work on, a goal outcome $\omega_g \in \Omega_T$ and a strategy $\sigma$ have been selected to optimize progress and according to an interest map (see below).
The selected strategy $\sigma$  applied to the chosen goal outcome $\omega_g$ chooses a sequence of controllables $l_c=  [c_1, \dots, c_m] $ as a candidate to reach the  $\omega_g$ (Alg.\ref{algorithm}, l.\ref{algo:sequence}). 

 {SGIM-SAHT chooses an adequate task $T_i$ , i.e. when the input space of $L_{Ti}$ includes $ (s,\omega_g)$. Then it} applies $L_{Ti}$ to find the action $a^i=L_{Ti}(c_i)$. This inference process may be recursive until the output is an action, using the hierarchy between tasks.  SGIM-SAHT thus infers from $l_c $  a compound action $a = [a_1, \dots, a_n] \in \mathcal{A}^{\mathbb{N}}$, to be executed by the robot. 
The trajectory of the episode with the primitive actions and controllables sequence and  {goal and reached outcomes $\omega_g,\omega_r$ are recorded in the memory}
 (Alg.\ref{algorithm}, l.\ref{algo:memory}).  
 Then, it computes the learner's competence for the goal outcome. In the RL framework, this competence can be seen as the reward for the goal outcome. In our multi-task learning setting, we use  {as competence a  reward function common to all goals} based on the Euclidean distance between the goal outcome $\omega_g$ and the reached outcome $\omega_r$ (Alg.\ref{algorithm}, l.\ref{algo:competence}). 
The memory and competence are used to update the models $M_T$ and $L_T$, the set of tasks, and the hierarchy of models $\mathcal{H}$ (Alg.\ref{algorithm}, l.\ref{algo:M}). These mechanisms differ in our 3 algorithms. 
Besides, the competence is used to obtain an interest map that associates to each strategy and region of outcome space partition an interest measure to guide the exploration. The interest measure is computed as the progress or derivative of the competence for the goal outcomes (see \citep{Nguyen2012PJBR}).

When the number of outcomes added to a region  $R_i$ exceeds a fixed limit, the region is split into two regions with a clustering boundary that separates outcomes with low from those with high interest (details in \citep{Nguyen2012PJBR}). This mechanism of progress-based intrinsic motivation leads to \textbf{automatic curriculum}, automatically organising the training phase from simple tasks, for which progress can be made earlier, to complex tasks, for which progress are made later. This curriculum is all the more structured as complex tasks are compositions of simpler tasks in the hierarchical learning setting.

We detail IM-PB and CHIME and compare in the next subsections. The description SGIM-PB will be detailed in the next chapter.

\subsection{Learn Task Hierarchy from a Static Set of Tasks}

\begin{figure}[t!]
\centering
\includegraphics[width=0.35\hsize]{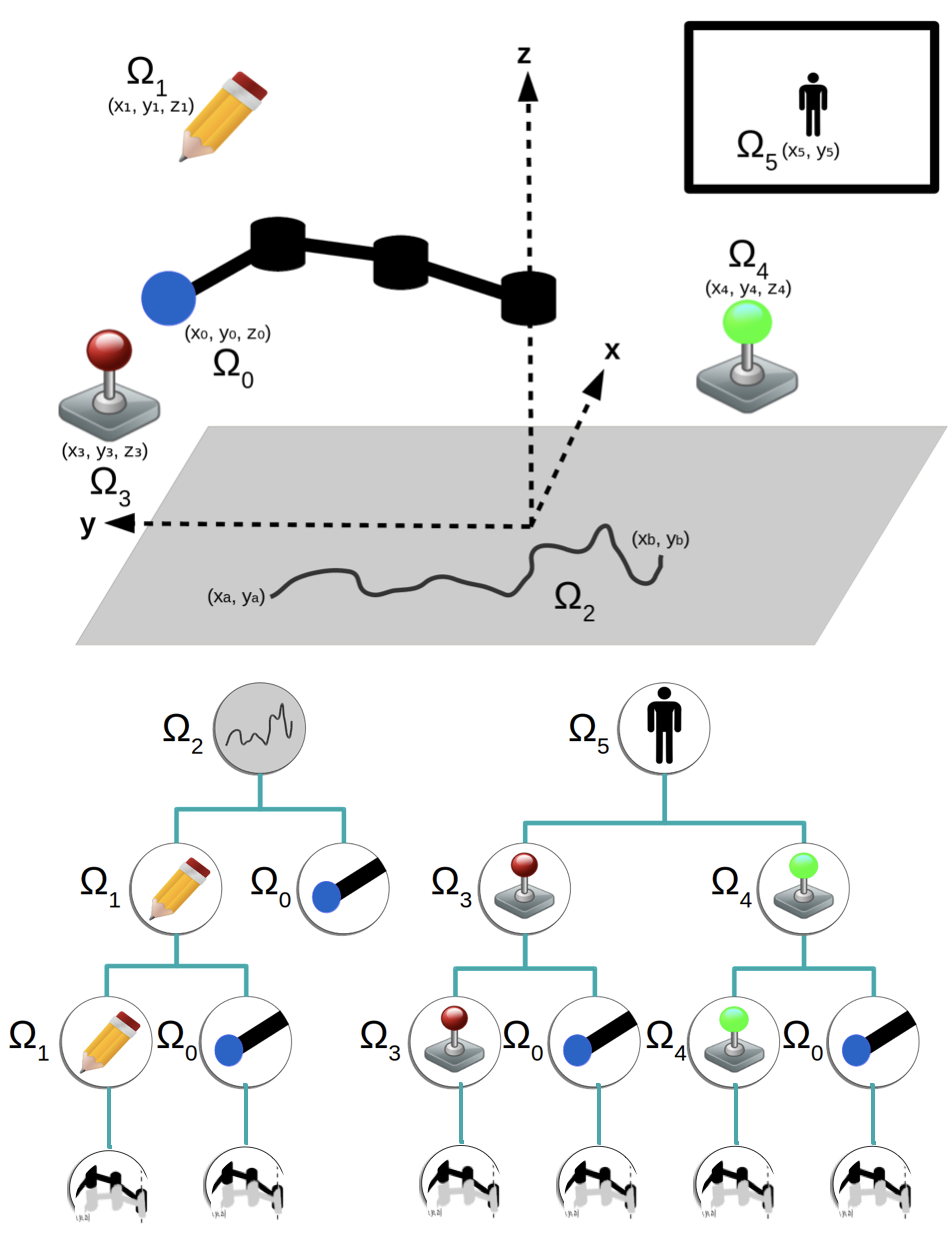}
\caption{Top: Experimental setup: a robotic arm, can interact with the different objects in its environment (a pen and two joysticks). Both joysticks enable to control a video-game character (represented in top-right corner). A grey floor limits its motions and can be drawn upon using the pen (a possible drawing is represented). Bottom: Task decomposition as designed in the experimental setup.
}
\label{fig:setupImpb}
\end{figure}

\begin{figure}[h!]
\centering
\includegraphics[width=0.25\linewidth]{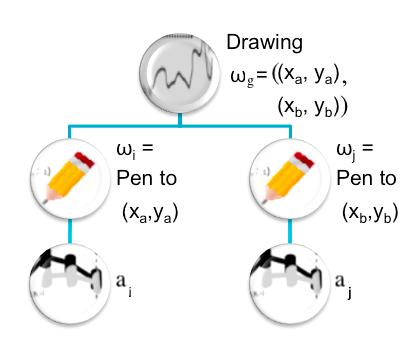}
\vspace{-0.3cm}
\caption{Representation of task hierarchy as a sequence of parametrised subgoals. To make a drawing $\omega_g$ between points $(x_a, y_a)$ and $(x_b, y_b)$, a robot can recruit subtasks consisting in ($\omega_i$) moving the pen to $(x_a, y_a)$, then ($\omega_j$) moving the pen to $(x_b, y_b)$. These subtasks will be completed respectively with  actions $a_i$  and $a_j$. To complete this drawing, the learning agent can use the sequence of actions $(a_i,a_j)$. $\omega_i, \omega_j, a_i, a_j$ are inferred by regression.
}
\label{fig:hierarchyPen}
\vspace{-0.6cm}
\end{figure}

A first stream of works proposed a representation of task hierarchy as a sequence of parametrised subgoals, in a teleological  approach \citep{Csibra2003PTRSLSBS} which considers {actions as goal-oriented}. As illustrated in Fig. \ref{fig:hierarchyPen}, a task is specified with its parameters and can be decomposed as  sequence of two subgoals, which are specified by variables and their parameters. This decomposition can be recursive, providing generalisability to several levels of hierarchy. Learning a task decomposition thus means identifying both subtasks and their parameters.

 Let us consider that the set of tasks is given. The robot needs to choose which are easier to learn first, which are more difficult, and which tasks can be reused as sub-goals for more difficult models: we use knowledge transfer in the context of automatic curriculum learning. To learn the hierarchy $\mathcal{H}$ between tasks, we proposed in  \citep{Duminy2018PIICRC} an implementation, IM-PB (Intrinsically Motivated Procedure Babbling) based on sequencing and interpolation of subgoals. 
Learning to draw with a pen and control a game character with joysticks, as illustrated in Fig. \ref{fig:setupImpb}, IM-PB uses intrinsic motivation to learn sequential tasks by composing more simple tasks by learning autonomously  from a known set of possible decompositions. IM-PB  involves an exploration of the possible task decompositions, driven by intrinsic motivation. 

The results in \citep{Duminy2018PIICRC} report that IM-PB can learn the task decomposition as designed, owing to the task hierarchy representation. IM-PB can then use the learned relationship between tasks to transfer knowledge from simple tasks to solve higher level tasks.

\subsection{Symbolic Representations of Tasks}
As I tackle the learning in continuous open-ended environments, the exploration of the continuous high-dimensional sensorimotor space faces the curse of dimensionality. In IM-PB, the curse of dimensionality problem increases with the  addition of the space of all possible task decompositions to the exploration. Our experimental setup features a task space  that is a composite and continuous space with tasks  up to 4 dimensions and the action space is of dimension 14. Our results show that SGIM-PB can handle such complexity to learn hierarchical tasks.  
However, scaling this algorithm to higher dimensional sensorimotor spaces can prove difficult. Thus, in order to improve the exploration of the task decomposition space, \textbf{as a means to reduce the size of the task space, I examined whether one can derive a symbolic (ie. discrete) representation of tasks or task decomposition}.

\subsection{Learn Task Hierarchy from a Dynamic Set of Tasks : Adaptive learning of an affordance hierarchy through exploration}

In the first line of work, I removed an assumption made in the previous studies with SGIM :  that the specification of all tasks is known from the beginning by the robot. Thus, the set of all possible task decompositions are known, and mappings only need their parameters to be tuned. 
In  \citep{Manoury2019PICHI}, on the contrary, the learning agent only discovers through its exploration which observables of the environment are controllable and creates a new task to learn. Thus, the robot learns which objects it can control. More concretely, it creates a new neural network to learn the mapping for this task, with the observable as output. To allow quick learning with few data, I considered how to use small size neural networks, the algorithm CHIME selects the appropriate inputs through an evolutionary selection. 
 The criteria for the stochastic selection is the estimated prediction accuracy.  Once the input and output of the neural network chosen, the weights are learned by supervised learning with the data collected during intrinsically motivated exploration. 
Moreover, a task A is added to the set of possible subtasks only when the robot reaches a mastery level for task A above a threshold. Thus, the space of all possible task decompositions grows progressively as the robot masters tasks. For instance, the manipulation of an object of the environment needs to reach a good level of mastery, before this object can be used as a tool to solve more complex tasks.

Thus, \textbf{CHIME implements the notion of tool use and affordance learning \citep{Gibson1979TheTheoryof} with a mechanism of emergence of tasks and a mechanism for task composition into hierarchically more complex tasks based on the same formulation of intrinsic motivation}.  While it uses deep neural networks, its originality lies in the proposition of a network which architecture evolves throughout its learning curriculum, enabling it to learn simple tasks quickly. The nested models emerge online, in parallel to the learning by the robot of its automatic curriculum.  The results in \citep{Manoury2019PICHI} show that the robot, owing to the intrinsic motivation heuristic, learns first the easy tasks, before the tasks higher in hierarchy: the reward in controlling its self-position increases first, in controlling the position of a movable object after, and at last pushing an object with another. 

Like SGIM-PB, CHIME implements a temporal abstraction representation of tasks : a task is not defined by its duration but by its goal.

\begin{figure}[h!]
    \centering
    \includegraphics[width=0.5\linewidth]{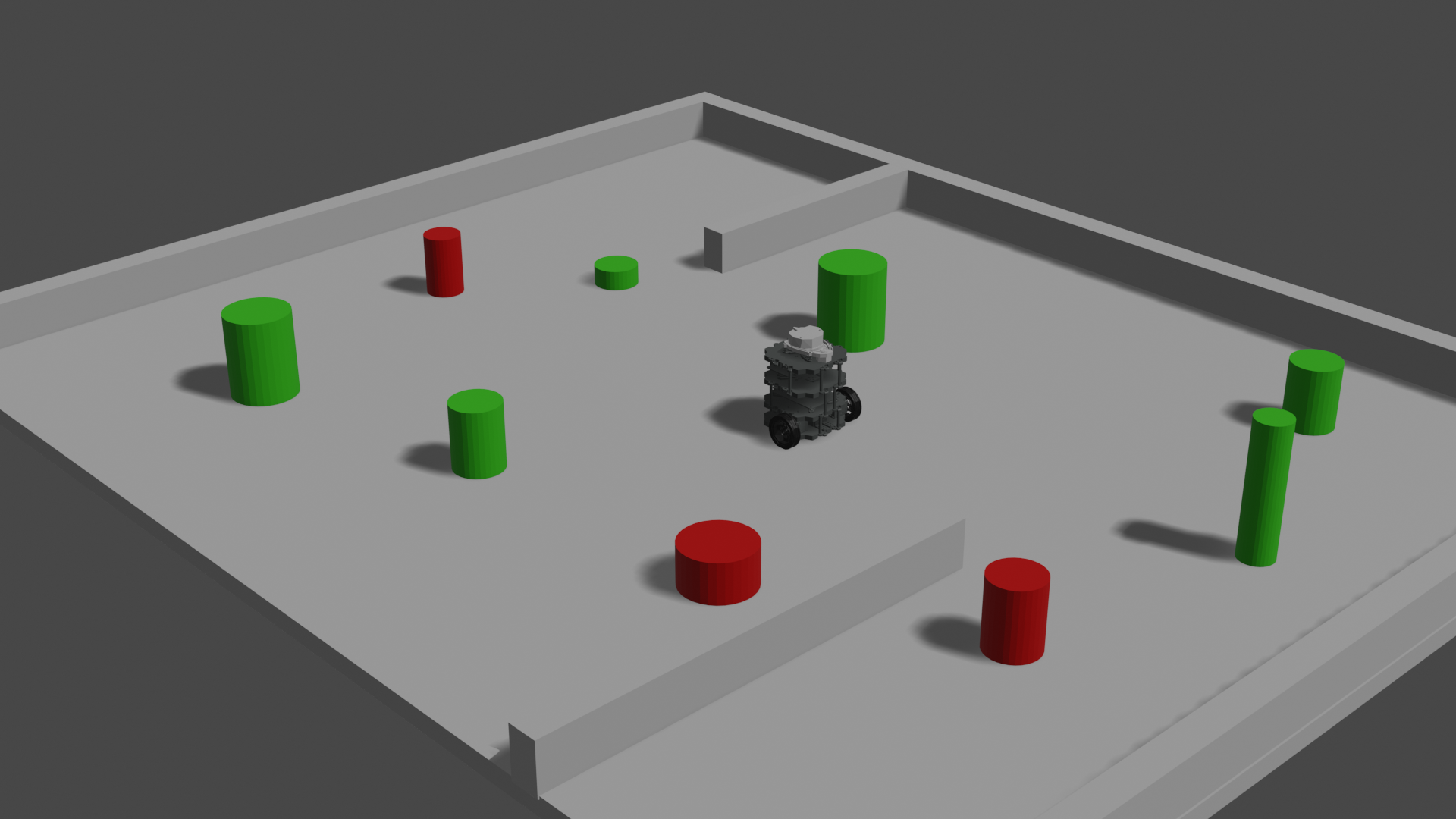}
    \caption{Experimental setup used: at the center is a mobile robot, the green objects represent movable entities, at the opposite of the red ones. The room is closed.}
    \label{fig:chimea:setup}
\end{figure}

The experimental setup used in this chapter is presented in Figure \ref{fig:chimea:setup}.

\section{STAR : Goal-Conditioned Hierarchical Learning}

Yet, CHIME still shows limitations for scaling to higher dimensional spaces because each task defines a continuous goal space. This is why in a second line of work, we build upon the framework of hierarchical reinforcement learning (HRL)\citep{Dayan1992ANIPS,Sutton1999AI}, which offers a potential solution for learning long-horizon tasks by training a hierarchy of policies. However, the abstraction used by policies is critical for good performance. Hard-coded abstractions often lack modeling flexibility and are task-specific.  Moreover, for long-horizon tasks, it is crucial to engage in hierarchical reasoning across spatial and temporal scales. This entails planning abstract subgoal sequences, thus we aimed for ab abstract representation of space. We examined how to discretise online during exploration the continuous goal space \hl{using formal methods, in collboration with Sergio Mover, from LIX, Ecole Polytechnique, expert in formal methods}. Recent studies show that representations that preserve temporally abstract environment dynamics are successful in solving difficult problems and provide theoretical guarantees for optimality.   In \citep{Zadem2024TICLR}, we propose a novel three-layer HRL algorithm, named Spatio-Temporal Abstraction via Reachability (STAR), that uses, at different levels of the hierarchy, both a spatial and a temporal goal abstraction. The goal space abstraction through reachability analysis results in a discretisation of the goal space, and thus a reduction of the exploration space. Thus, \textbf{STAR  exploits an emergent discrete representation of the goal space, to be used more easily for task composition into sequential tasks}. 
 
 STAR  uses as framework, a goal-conditioned Markov Decision Process $(\states, \mathcal{A}, P, r_{ext})$, where $\states \subseteq \real^n$ is a continuous state space, $\mathcal{A}$ is an 
action space, $P(s_{t+1} \lvert s_t, a_t)$ is a probabilistic transition function, and $r_{ext} : \states \times \states \rightarrow \real$ is a parameterised reward function, defined as the negative distance to the task goal $\taskgoal \in \states$, i.e $r_{\text{ext}}(s,g^*) = - \dist{\taskgoal - s}$.
The \emph{multi-task reinforcement learning problem} consists in learning a goal conditioned policy $\pi$ to sample at each time step $t$ an action $a \sim \pi(s_t \mid \taskgoal)$, so as to  maximize the expected cumulative reward. The spatial goal abstraction is modeled by a \emph{set-based abstraction} defined by a function $\abs: \states \rightarrow  2^\states$ that
maps concrete states to sets of states (i.e.,  $ \forall s \in \states,  \abs(s) \subseteq \states$).
We write $\partitions_\abs$ to refer to the range of the abstraction $\abs$, which is intuitively the abstract goal space. We further drop the subscript (i.e., write $\partitions$) when $\abs$ is clear from the context and denote elements of $\partitions$ with the upper case letter $\partition$.

\begin{figure}[tb]
  \centering
  \includegraphics[width=\textwidth]{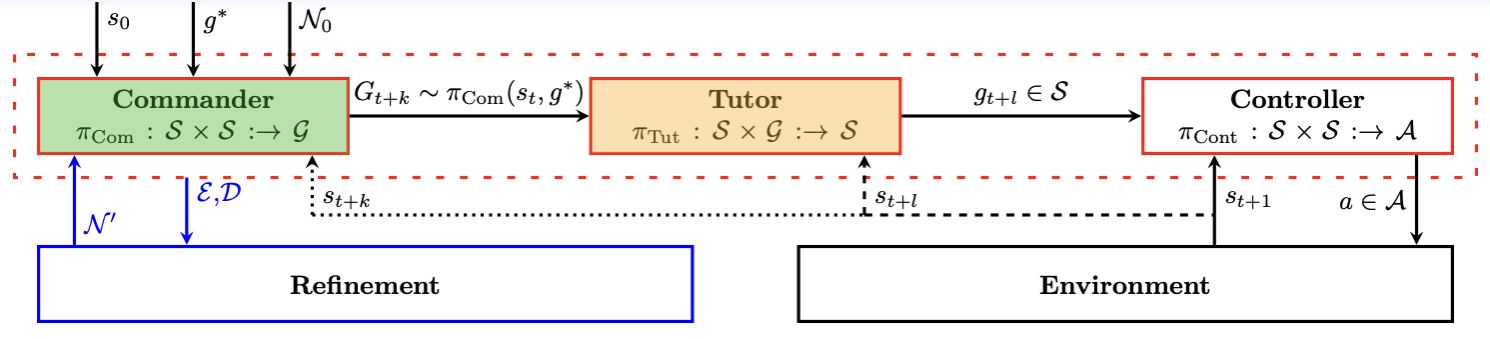}
  \caption{\textbf{Architecture of STAR}.
  The algorithm's inputs are the initial state $s_0$, the task goal $g^*$, and an initial abstraction $\abs_0$.
  STAR runs in a feedback loop a Feudal HRL algorithm (\textcolor{red}
  {dashed red} block) and an abstraction refinement (\textcolor{blue}
  {blue} box). 
  The \textcolor{red}
  {solid red} blocks show the  HRL agents (Commander, Tutor, Controller). 
  The agents run at different timescales ($k > l > 1$), shown with the solid, dashed, and dotted lines carrying the feedback from the environment to the agents.
  The Refinement uses as inputs the past episodes ($\mathcal{D}$) and a the list of abstract goals ($\edges$) visited during the last episode, and outputs an abstraction.
  }
  \label{fig:architectureStar}
\end{figure}

STAR learns, at the same time, a spatial goal abstraction $\abs$ and policies at multiple time scales. 
The STAR algorithm, shown in Figure~\ref{fig:architectureStar}, has two main components: a 3-levels Feudal HRL algorithm (enclosed in the \textcolor{red}{red dashed lines}); and an abstraction refinement component (shown in the \textcolor{blue}{blue solid lines}).
STAR runs the Feudal HRL algorithm and the abstraction refinement in a feedback loop, refining the abstraction $\abs$ at the end of every learning episode.

Our work presents the following contributions:
\begin{enumerate}
\item A novel Feudal HRL algorithm, STAR, to learn online 
 
 (Fig. \ref{fig:architectureStar}) with the three RL agents : 
\begin{enumerate}
    \item \navigator: the highest-level agent learns the policy $\navpolicy: \states \times \states \rightarrow \partitions$ that is a goal-conditioned on $\taskgoal$ and samples an abstract \textbf{goal} $\partition \in \partitions$ that should help to reach the task goal \taskgoal~ from the current agent's state ($\partition_{t+k} \sim \navpolicy(s_t, \taskgoal)$).
    \item \manager: the mid-level agent is conditioned by the \navigator goal $G$. It learns the policy $\manpolicy: \states \times \partitions \rightarrow \states$ and picks \textbf{subgoals} in the state space 
    ($g_{t+l} \sim \manpolicy(s_t,\partition_{t+k}) $).
    \item \controller: the low-level policy $\contpolicy: \states \times \states \rightarrow \mathcal{A}$ is goal-conditioned by the \manager's subgoal $g$ and samples actions to reach given goal ($a \sim \contpolicy(s_t, g_{t+l})$).
    \end{enumerate}

\item An emergent symbolic representation of the environment, based on reachability analysis. Each symbol representing a region of the state space, we estimate the k-step reachability of the forward model from each region. We refine the representation by splitting the regions online. We provide a theoretical motivation for using reachability-aware goal representations, showing  convergence to a reachability-aware abstraction after applying a finite number of refinements, and showing  a guarantee of a suboptimality bound for the converged policy. 

Compared to the optimal policy, the policy learned on the abstraction has a near optimal value : $| V_{\pi*}(s_i) - V_{\pi*_{\mathcal{N}}}| \leq U(\epsilon,i,\gamma)|$, where U is a function of $\gamma$  the discount factor, $\epsilon$ a bound on reward values, and $i$  the time step index. 

\item Empirical results showing that STAR successfully combines both temporal and spatial abstraction for more efficient learning, and that the reachability-aware abstraction scales to tasks with more complex dynamics in Ant environments (Fig. \ref{fig:ant_envs}).
\end{enumerate}

\newcommand{\mazewidth}{0.25} 
\begin{figure}[tbh]
  \begin{center}
    \subfloat[{Ant Maze}\label{fig:ant_maze}]{%
      \includegraphics[width=\mazewidth\columnwidth]{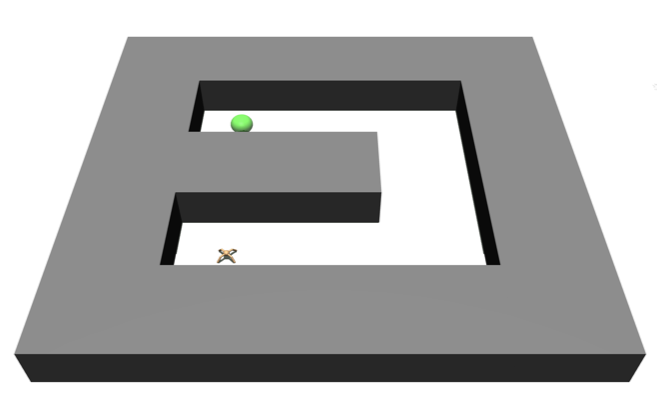}
    }
    \subfloat[{Ant Fall}\label{fig:ant_fall}]{%
      \includegraphics[width=\mazewidth\columnwidth]{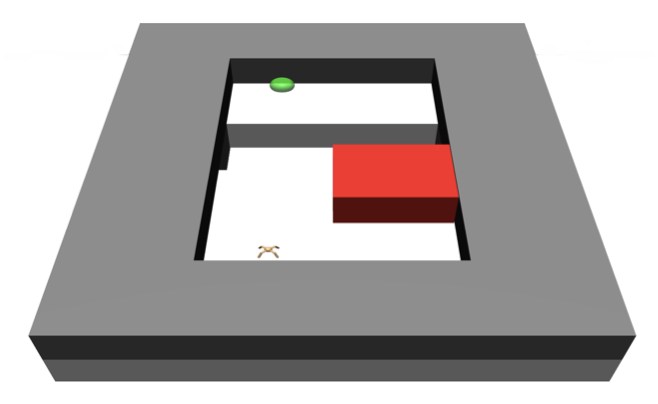}
    }
    \subfloat[{Ant Maze Cam}\label{fig:ant_maze_cam}]{%
      \includegraphics[width=\mazewidth\columnwidth]{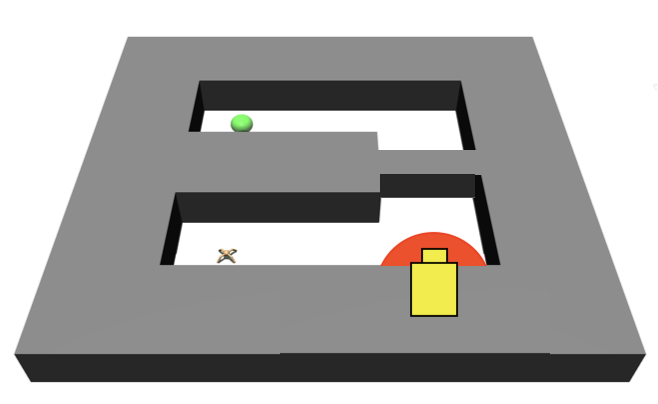}
    }
  \end{center}
  \caption{Ant environments. \textbf{Ant Maze:} in this task, the ant must navigate a '$\supset$'-shaped maze to reach the exit positioned at the top left.
     \textbf{Ant Fall:} the environment is composed of two raised platforms seperated by a chasm. The ant starts on one of the platforms and must safely cross to the exit without falling. A movable block can be push into the chasm to serve as a bridge. Besides the precise maneuvers required by the ant, falling into the chasm is a very likely yet irreversible mistake.
     \textbf{Ant Maze Cam:} this is a more challenging version of Ant Maze. The upper half of the maze is fully blocked by an additional obstacle that can only be opened when the ant looks at the camera (in yellow) when on the red spot. The exit remains unchanged. }
  \label{fig:ant_envs}
\end{figure}

\begin{figure}[tbh]
    \centering
    \includegraphics[width=0.7\columnwidth]{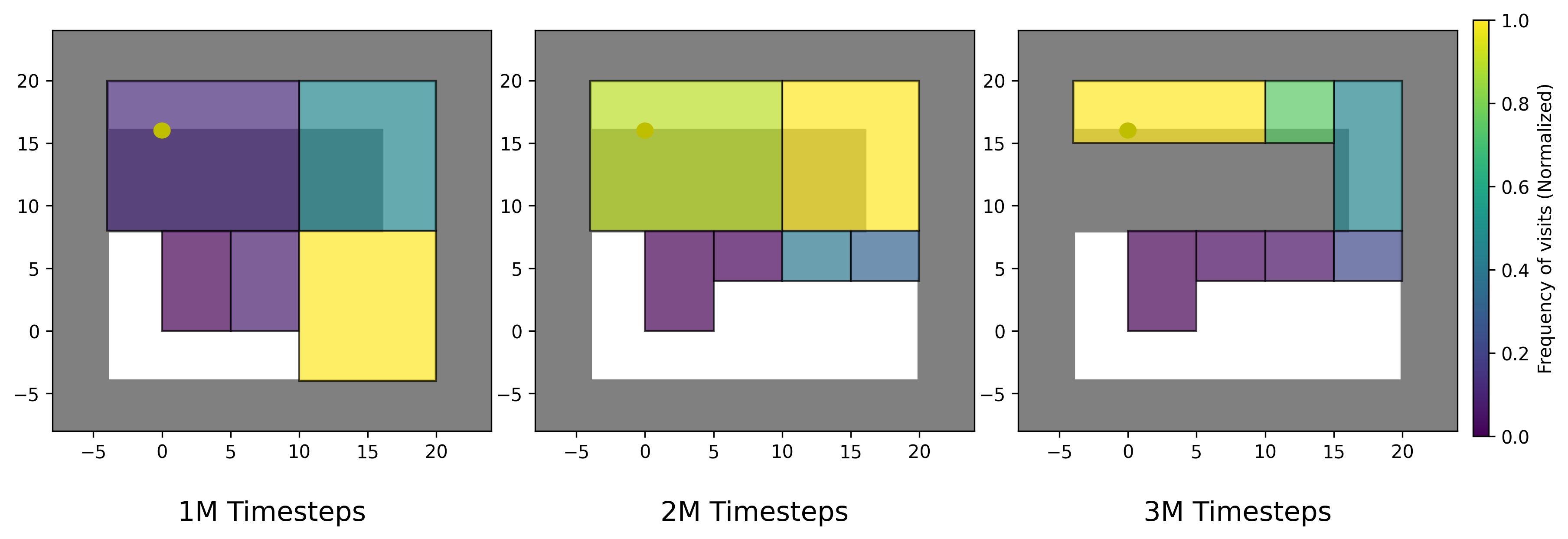}
    \caption{Frequency of goals visited by the \navigator when evaluating a policy learned  after 1M, 2M, and 3M timesteps (averaged over 5 different evaluations with 500 maximum timesteps).
    The subdivision of the mazes represent (abstract) goals.
    The color gradient represents the frequency of visits of each goal.
    Grey areas correspond to the obstacles of the environment in Ant Maze.
    }
    \label{fig:antmaze_repr}
\end{figure}

Our evaluation of the Ant environments show superior success rate of STAR compared to the state of the art. 
To analyse how the learned abstraction decomposes the environment to allow for learning successful policies, we examine the abstraction learned by STAR's \navigator agent at different timesteps during learning when solving the Ant Maze.
Fig. \ref{fig:antmaze_repr} provides some insight on how STAR 
gradually refines the goal abstraction to identify successful trajectories in the environment. We can see that, progressively, the ant explores trajectories leading to the goal of the task. The most visited areas move from regions before the obstsacle wall to the corner past the wall, then finally the region of the goals. Additionally, the frequency of visiting  goals in the difficult areas of the maze (e.g., the tight corners) is higher, and these goals are eventually refined later in the training, fitting with the configuration of the obstacles. 
 In particular, STAR learns a more precise abstraction in bottleneck areas where only a few subset of states manage to reach the next goal.

\section{Representations used for the Analysis of Human Activity of Daily Living}
In parallel to the previous investigations on representations of sequential tasks for learning agents to build them up, we also examined from the observational point of view, representations used in analysis methods. 
As the most complex motor activities in terms of compositionality and hierarchical decomposition into subtasks, we inspected our daily tasks : \emph{activities of daily living} (ADL) include sleeping, eating, cooking, bathing ... as listed in datasets such as CASAS \citep{Cook2012C} or the ADL definitions provided by  \citet{Katz1983JAGS}.   Indeed, an activity of daily living such as cooking and cleaning can vary greatly from one day to the other depending on the context and goal of the inhabitant. Moreover, ADL are sequential actions of variable length combining several primitive actions with multilevel temporal dependencies.  
Activities can also be deeply interconnected, underscoring the significance of temporal context in human activity recognition. Each activity can be viewed as a combination of unit actions that are selected and organised for the completion of a temporally distant goal. The activities of daily living share common properties as the sequential actions we considered earlier : the activities can be described as sequences of subtasks as multilevel time dependencies, the subtasks are recruited to complete a goal, and the temporal length and complexity degree of each task is not a-priori bounded. How complex actions such as activities of daily living can be described by hierarchical models ? 

Our investigations first examined the recognition of activities based on video data \hl{as part of the project AMUSAAL}. We proposed a hierarchical LSTM to classify the human poses in \citep{Devanne2019ICSC}, and showed that a \textbf{hierarchical representation is better} suited for activities of daily living. 

\hl{As part of a collaboration with specialists in smart homes, Christophe Lohr from IMT Atlantique and the company Delta Dore, w}e then examined the problem of recognition of activities based IoT network data with sensors motion, door open/close, temperature... Due to specificities of the privacy-respecting design even-triggered sensors, the time series to analyse does not have the Markov property. Understanding one sensor activation often requires contextualization from the history and from other sensors, as shown in \citep{Bouchabou2021E} where Word2Vec encoding can capture contextualized semantics, and in \citep{Bouchabou2023MLITSWEP} where  GPT transformer decoder can capture longer contexts.   Compounding these issues, ambient sensor data manifest as noisy, multi-variate, and irregular time series with long-term dependencies, hardly modelled by traditional machine learning and deep learning methods.  Examining multi-level temporal dependency, our work integrates (1) attention mechanisms for discerning the importance of sensor signals across a whole sequence , (2) pre-trained generative transformer embeddings capturing sensor inter-relations \citep{Bouchabou2023MLITSWEP}, (3) a hierarchical model emphasising activity succession for long-horizon dependency, and (4) a temporal encoding model to harness the timing of events. 
Despite semantic contextualisation, long-term dependency sequence models, and temporal encoding, a \textbf{hierarchical model of the time series leads to better recognition of the sequential activities}.  

Indeed, for human activity recognition, hierarchical models have been proposed as ontologies of context-aware activities for recognition of activities of daily living in smart homes \citep{Hong2009PMC}, hierarchical hidden Markov models \citep{Asghari2019} for recognition of activities of daily living in smart homes, or hierarchical LSTM with two hidden layers for activity recognition from wearable sensors \citep{Wang2020CSSP}. The hypothesis that sequential actions need to be represented by hierarchical models has been proposed in reinforcement learning : to solve sequential tasks, hierarchical reinforcement learning  \citep{Barto2003DEDS,Barto2013CRMHOB} enabled tackling sequential tasks by decomposing into subtasks. These machine learning models confirm the neuroscience description that our nervous system selects and organizes actions in a hierarchical model \citep{Grafton2007HMS}, as well as the behavioural psychology studies showing that humans use hierarchical representations of action sequences of efficient planning and flexibility \citep{Eckstein2021CCCSS}.  Our study provides  another computation model supporting this hypothesis, but has the particularity of analysing activities of daily living which are very complex and variable tasks.

In relation to our robot learning algorithms STAR, CHIME and IM-PB, these works seem to \textbf{support our hypothesis that robots building up sequential tasks should rely on a hierarchical representation of tasks}.

\chapter{Intrinsic Motivation to Learn with Teachers}

While in the previous chapter, the learning agent was exploring its environment  autonomously, in this chapter, I integrate social guidance in the learning process. 
 I used theories of imitation learning and intrinsic motivation to bridge the paradigms of reinforcement learning and supervised learning, and devise the framework of socially guided intrinsic motivation. I proposed algorithms in the framework of reinforcement learning to enable learning agents to \textbf{learn multiple parametrised tasks by  devising their own learning strategy: they choose actively what do learn, when to learn and thus their own curriculum; and also what, when and whom to imitate}.

From the point of view of cognitive science, our work can be seen as a first outline for model of the intrinsic motivation to interact with teachers, and thus a first step towards a \textbf{model of motivations for social interaction}.

\section{Learning a Parametrised Task}

My PhD research has taken a stance opposite to the current research in intrinsic motivation.
While computational models of intrinsic motivation have examined how  agents learn by interacting autonomously with its environment to control it without external reward by implementing an intrinsic reward based on its empirical progress, my PhD research has proposed a formulation of the intrinsic motivation of interacting with teachers: \textbf{what can drive a learning robot to actively interact with teachers?} Our proposition is inspired by developmental psychology studies that describe how children can learn by trial and error during their exploration with the environment, and also by requesting their parents' help in an active manner.

My  first step was to study how control learning in continuous sensorimotor spaces for parametrised tasks with a continuous goal space, can benefit from using both paradigms.

\begin{figure}[tb]
\centering
\includegraphics[width=0.6\textwidth]{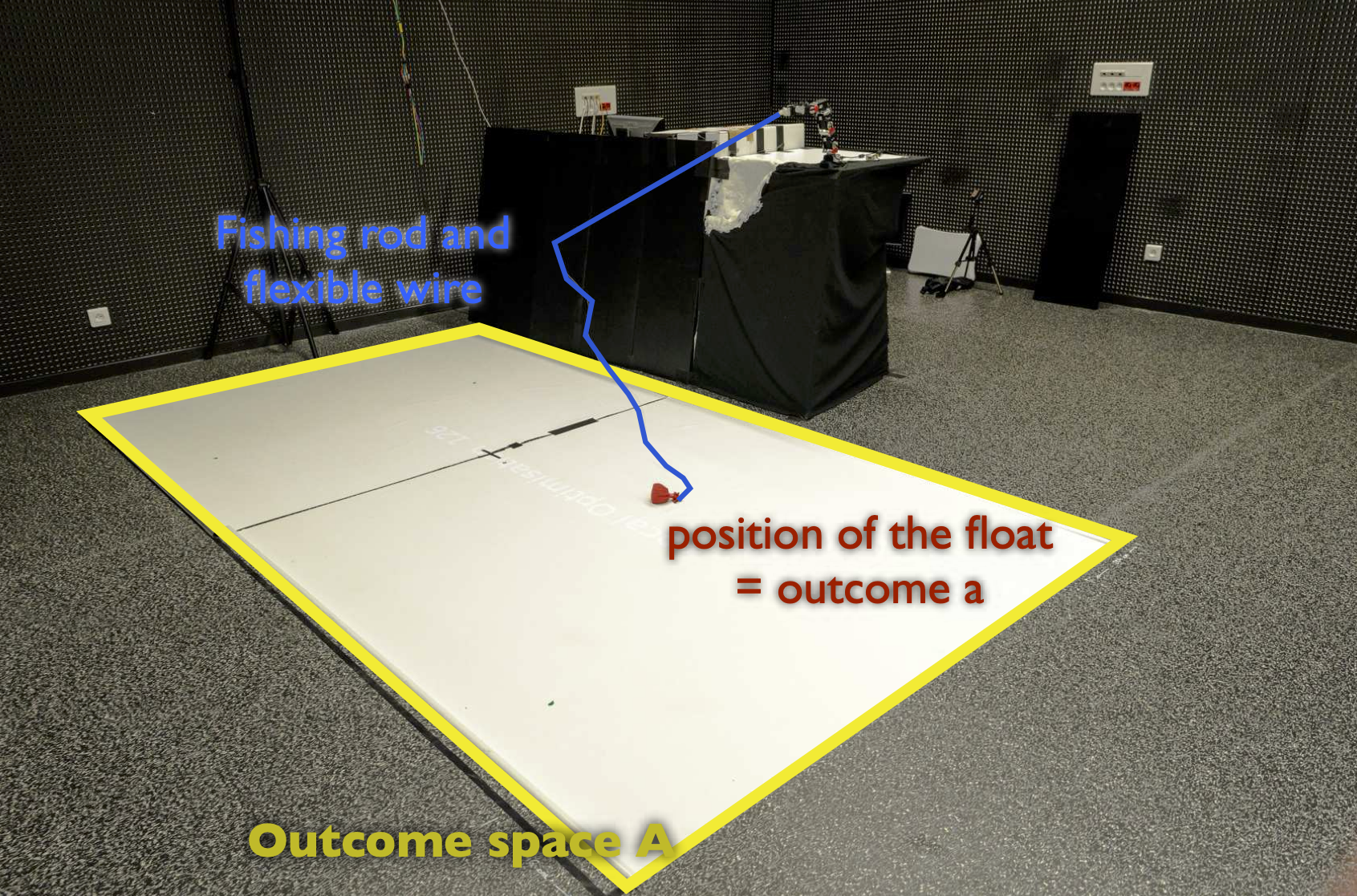}
\caption{
The robot observes the outcome $a$, the final position of the ball after its movement. It learns which movement can reach different positions on the floor. The camera is placed above the white surface, and can only see the white surface which materialises the outcome space $A$.  }
\label{fig:nofishField}
\end{figure}

\begin{figure}[tb]
  \centering
  \includegraphics[width=\textwidth]{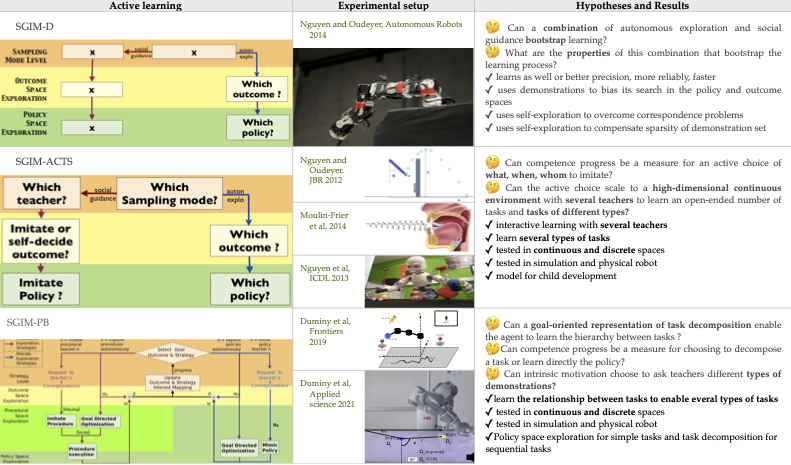}
\caption{Three implementations of the SGIM framework are presented with various illustrative experiments and their corresponding publication, as well as the main hypotheses and results. Each algorithmic architecture allows the agent to take active control of various aspects of its learning strategy. Each of the results presented allow us to present aspects of the advantages for a fully active system, that can decide on all aspects of its learning strategy.}
\label{fig:differentSGIM}
\end{figure}

\begin{figure}
\centering
\includegraphics[width= 0.9\textwidth]{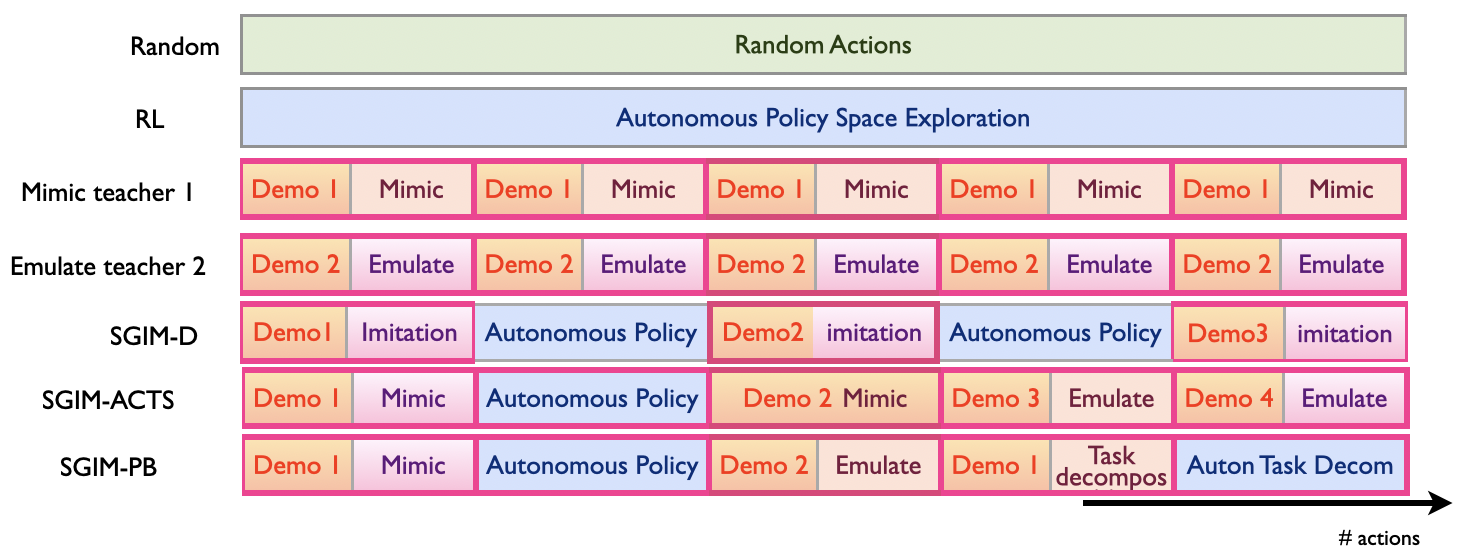}
\caption{ Comparison of the timeline of several learning algorithms : (a) Random exploration of the action space A; (b) autonomous exploration with Reinforcement Learning; (c) Mimicry to reproduce a demonstrated action, received at a fixed frequency; (d) Emulation learning to reproduce a demonstrated outcome, received at a fixed frequency; and (e) SGIM-D that requests demonstrations on a fixed frequency; (f) SGIM-ACTS that decides when and who to request demonstrations of actions and outcomes and (g) SGIM-PB that decides when and who to request demonstrations of actions or decomposition into subgoals. 
}
\label{fig:ExperimentalProtocol}
\end{figure}

With a robot arm learning policies for manipulating a fishing rod to place the float to any position (see fig \ref{fig:nofishField}), I showed that action and outcome demonstrations  by humans lead to better generalisation to a field of tasks, in \cite{Nguyen2014AR}. In \cite{Nguyen20122IISRHIC},  I showed that human demonstrations given by kinesthetics display different distribution characteristics in the trajectory profile. Comparing the demonstrations recorded from a human and movements provided by a robot that has finished learning, the most noticeable difference is that the joint value trajectories of the human demonstrator are all monotonous, and seem to have the same shape, only scaled to match different final values.  
Indeed, the comparison  indicates that  human demonstrations are not randomly generated but are well structured and regular, while the demonstrations from expert robots come from different probability distributions.
Therefore, the human demonstrator shows a bias through his demonstrations to the robot, and orients the exploration towards different subspaces of the policy space. Because \textbf{human demonstrations have the same shape, they belong to a smaller, denser and more structured subset of trajectories from which it is easier for the learner to generalise, and build upon further knowledge}.
 Thus, a robot can exploit this bias to learn to generalise to more tasks by using data from both autonomous goal-directed exploration and from a few demonstrations. Devising the framework of Socially Guided Intrinsic Motivation, I proposed several implementations, which are  particular cases of SGIM-SAHT (Algo. \ref{algorithm}).  Fig. \ref{fig:differentSGIM} outlines the difference between the algorithms, the experimental setups they were tested on, and the main results;  Fig. \ref{fig:ExperimentalProtocol} illustrates different strategies used by each algorithm on the timeline of the learning process.

First, the algorithm SGIM-D, for learning online parametrised tasks in a passive manner,  alternates at a fixed frequency between autonomous exploration and imitation learning \citep{Nguyen2014AR}. SGIM-D can produce more varied outcomes in the environment, covering a larger area of the task space, and is robust to large size task spaces, even when the set of reachable tasks covers a small region. This first  study \textbf{validates the benefits of combining both learning paradigms}, and allowed us to consider more varied interactions. 

Second, by assessing the progress thanks to each demonstration, I proposed a formulation of the intrinsic motivation to request for demonstrations from a teacher, and proposed the algorithm SGIM-ACTS to \textbf{actively collect data by choosing when to explore autonomously and when to request a demonstration; in which case, it chooses what to imitate (what type of demonstration to ask : a goal for emulation or an action for mimicry),
 and whom to imitate by learning each teacher's field of expertise} 
  \citep{Nguyen2012PJBR}. Thus, SGIM-ACTS can choose the learning strategy and its curriculum. It has been successful at learning parametrised control tasks of different nature from multiple teachers \citep{Nguyen2012PJBR}, at recognising 3D objects by collecting data through different manipulation strategies and by requesting the help of a human \citep{Ivaldi2013TAMD}.  
  
\section{Influence of demonstrations}

 We could also model the \textbf{influence of the mother tongue in the development of vocalisation in infants}, \hl{in a collaboration with Clement Moulin-Frier from Flowers Team, Inria} \citep{Moulin-Frier2014FP} : we simulated how an agent can activate the different muscles of the mouth and vocal tract to learn to produce sounds. Our findings show that the sounds produced during the automatic curriculum bears resemblance with child development described in behavioural psychology, but also that the influence of the mother tongue on the sounds produced can be explained by our formulation of intrinsic motivation to imitate the mother tongue's sounds.
Thus our formulation of intrinsic motivation for interacting with teachers enables better performance for learning robots from an engineering perspective, but also an element of \textbf{modelling child development in cognitive science}.

\section{Robustness to the quality of demonstrations}

I showed in these works that the SGIM algorithms is \textbf{robust to poor quality demonstrations}, comparing a helpful teacher who provides with different demonstrations with a teacher  who repeats the same demonstrations from a small demonstration dataset, we show in \citep{Nguyen2013IICDLE} that the robot's performance is not impacted, contrarily to a random choice of strategies : it can choose the right learning strategy to cope with poor quality demonstrations. 
SGIM can handle correspondence problems that occur when the robot can not move like the teacher, due for instance to correspondance problems, which come from differences in their bodies \citep{Nguyen2012PJBR,Duminy2016I2JIICDLER}.

\section{Compositional Tasks with Active Imitation Learning}

With the algorithm SGIM-PB (Socially Guided Intrinsic Motivation with Procedure Babbling), \citet{Duminy2021AS} extends, this intrinsic motivation criteria for requesting from the best teacher either a demonstration of a policy or a task decomposition. It uses the same representation for task decomposition as IM-PB, presented in the previous chapter. The interaction with teachers thus has a supplementary strategy: asking how to decompose a task. SGIM-PB  outperforms  SGIM-ACTS on the most complex outcome spaces, owing to task decomposition which enables it to learn and exploit the task hierarchy of this experimental setup and previously learned skills.

\begin{figure}[H]
\includegraphics[width=0.7\linewidth]{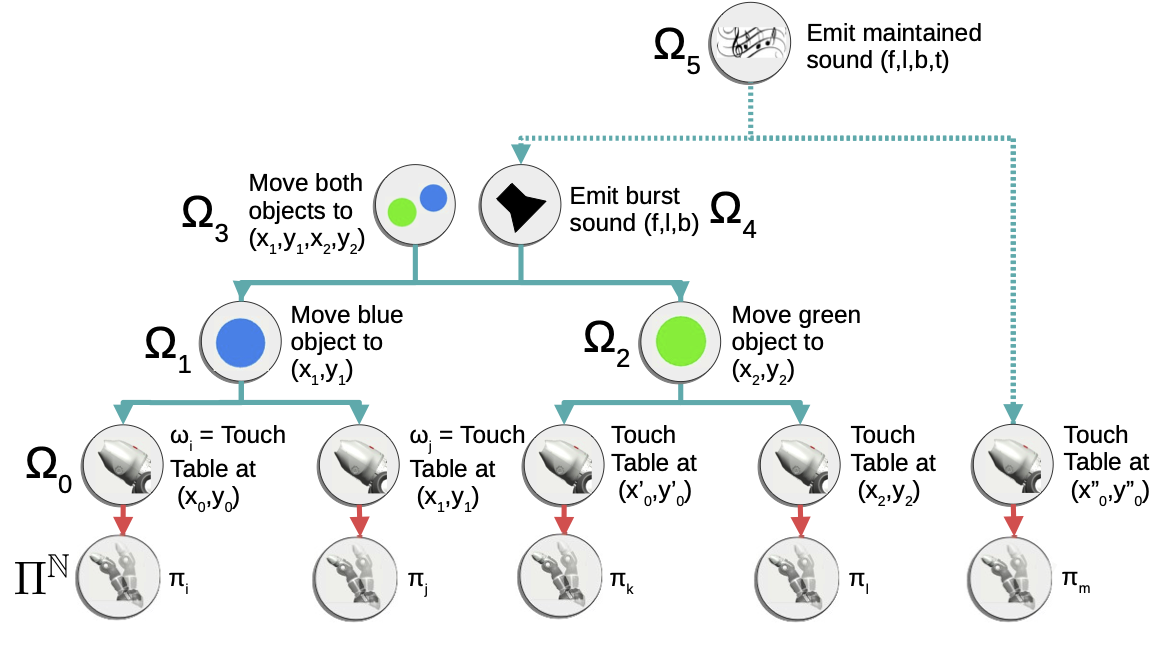}
\caption{Task hierarchy of the Yumi experimental setup : blue lines represent task decomposition  for the simulation setup, dashed lines for the physical setup,  red lines for the direct inverse model. Both task decomposition and inverse model are learned online. Eg. to move a blue object placed initially at $(x_0,y_0)$ to a desired position $(x_1,y_1)$, the robot can carry out task ($\omega_i$) that consists in moving its end-effector to $(x_0, y_0)$ to pick it, then task ($\omega_j$) that consists in moving the end-effector to $(x_1, y_1)$. These subtasks are executed  with respectively action primitives $\pi^i$  and $\pi^j$. Therefore to move the object, the learning agent can use the sequence of action primitives $(\pi^i,\pi^j)$. }
\label{fig:hierarchy}
\end{figure}

For a 6 DOF robot arm learning hierarchical tasks as represented in Fig. \ref{fig:hierarchy}, our results also show that demonstrations of task decomposition can efficiently replace demonstrations of sequential motor policy. Most of all, our robot \textbf{learned to ask  demonstrations of motor policies for simple tasks and to ask  demonstrations of task decomposition for sequential tasks} \citep{Duminy2018ICSC}. Likewise, for autonomous exploration, it explores directly the policy space for simple tasks and the task decomposition space for sequential tasks. These behaviours derive from the same criteria for active learning : the learning progress. This intrinsic reward is thus a common criteria for choosing tasks and for choosing teachers.

\begin{figure}[H]
\center
\includegraphics[width=0.6\linewidth]{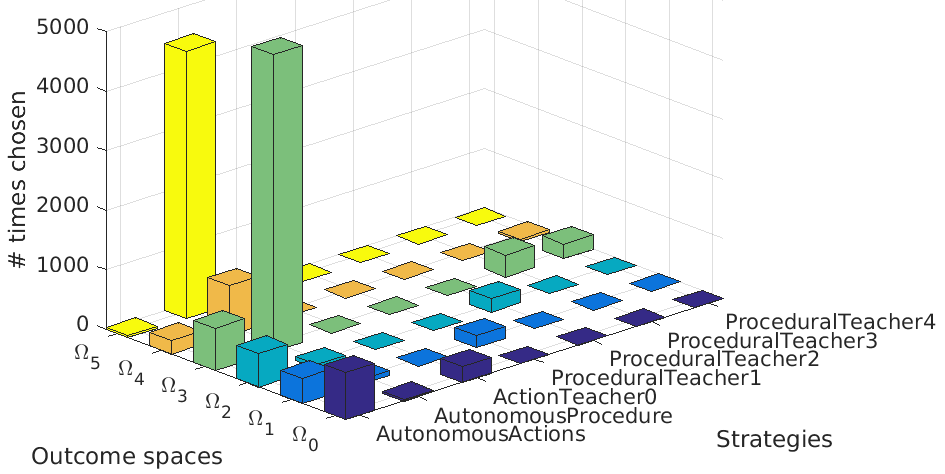}
\caption{Choices of strategy and goal outcome for the SGIM-PB learner during the learning process. Strategies are autonomous exploration of the action  space, task decomposition (=procedure), mimicry of actions, or imitation of task decomposition (=procedural teacher). }
\label{fig:physical_task_strategy}
\end{figure}

To understand the reasons for the better performance of SGIM-PB on a physical robot setup, let us examine the histogram of the strategies chosen per task in Figure~\ref{fig:physical_task_strategy}. For the most complex task, $\Omega_5$, SGIM-PB uses massively task decomposition space exploration(AutonomousProcedure) compared to action space exploration (Autonomous Actins). Owing to autonomous task decomposition exploration, SGIM-PB can thus learn complex tasks by using the decomposition of tasks into known subtasks. This highlights the \textbf{essential role of the task decomposition representation and the task decomposition space exploration by imitation but also by autonomous exploration, in order to learn high-level hierarchy tasks, which have sparse rewards}, so as to {adapt the complexity of the action sequences to the hierarchy level of the goal task}.

\begin{figure}[H]
\includegraphics[width=0.8\linewidth]{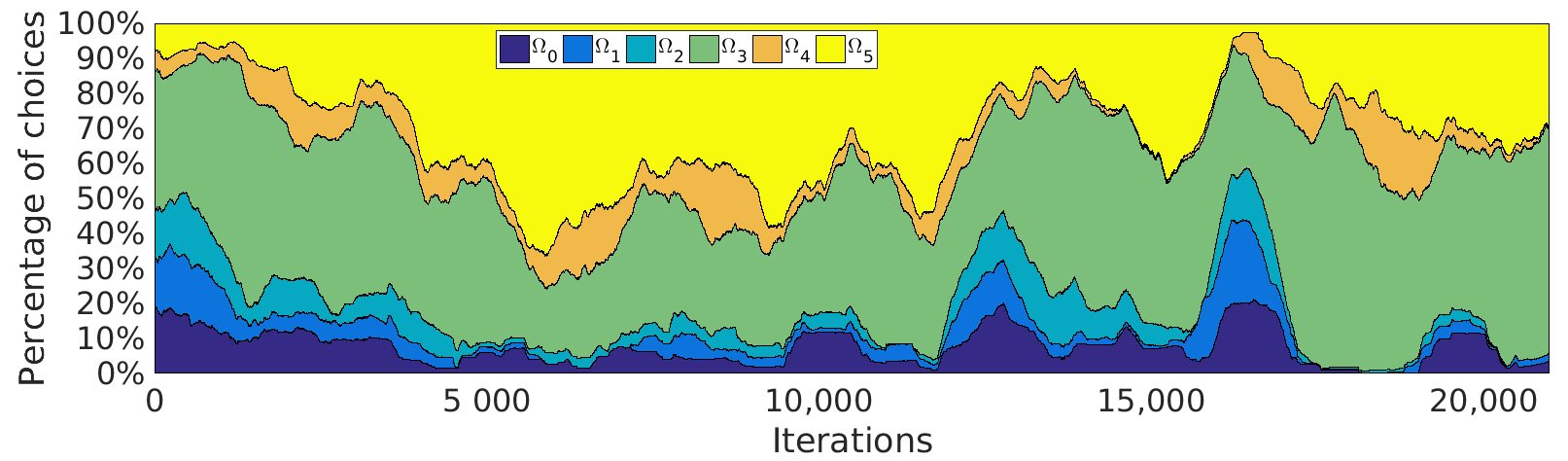}
\caption{{Automatic Curriculum Learning : Evolution}   of choices of tasks for the SGIM-PB learner during the  learning process on the physical setup.}
\label{fig:physical_tasks}
\end{figure}

\begin{figure}[H]
\includegraphics[width=0.8\linewidth]{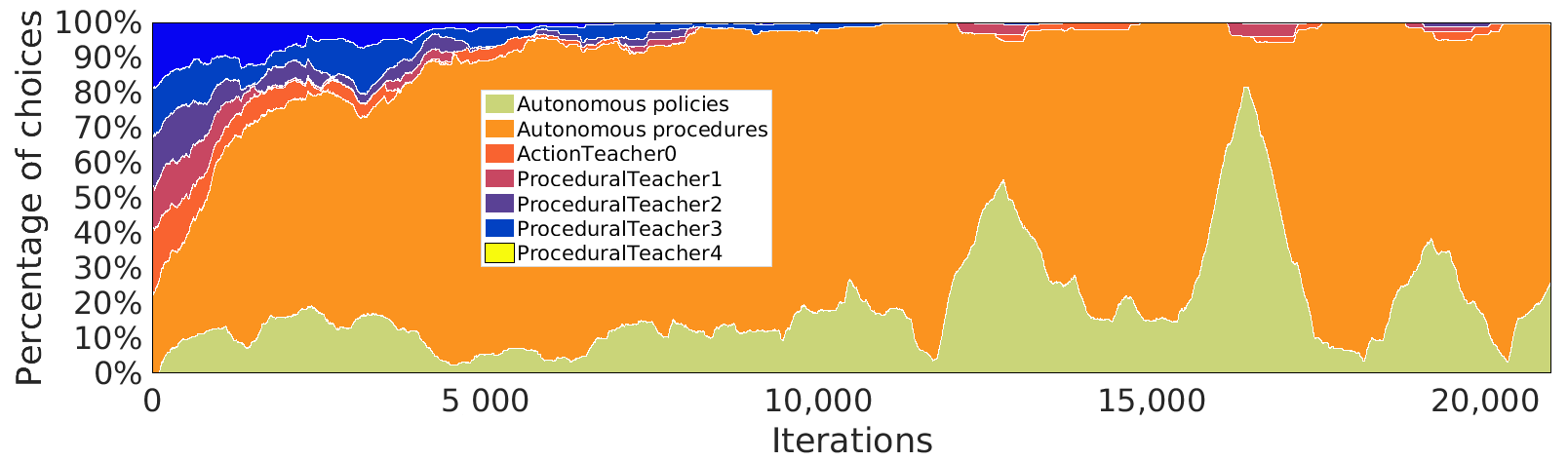}
\caption{{Evolution}  
 of choices of strategies for the SGIM-PB learner during the  learning process on the physical setup.}
\label{fig:physical_strats}
\end{figure}

Besides, the evolution of outcome spaces (Figure~\ref{fig:physical_tasks}) and strategies (Figure~\ref{fig:physical_strats}) chosen during training show the same trend from iterations 0 to 10,000: the easy outcome spaces  $\Omega_0, \Omega_1, \Omega_2$ are more explored in the beginning before being neglected after 1000 iterations, to explore the most complex outcome spaces $\Omega_3, \Omega_5$.  
Imitation is mostly used in the beginning of the learning process, whereas later in the automatic curriculum, autonomous exploration is preferred. 

\vspace{-6pt}

To summarise, SGIM-PB shows the following characteristics :

\begin{itemize}

	\item Hierarchical RL: it learns online task decomposition on 4 levels of hierarchy using the procedural framework; and it exploits the task decomposition to match the complexity of the sequences of action primitives to the task;
	\item Automatic curriculum learning: it autonomously switches from simple to complex tasks, and from exploration of actions for simple tasks to exploration of procedures for the most complex tasks;
	\item Imitation learning: it empirically infers which kind of information is useful for each kind of task and requests just a small amount of demonstrations with the right type of information by choosing between procedural and action teachers;
	\item Transfer of knowledge: it automatically decides what information, how, when to transfer and which source of information for cross-task and cross-learner transfer learning.
\end{itemize}

  Thus, \textbf{our work proposes an active imitation learning algorithm based on intrinsic motivation that uses empirical measures of competence progress to choose at the same time what target task to focus on, which source tasks to reuse and how to transfer knowledge about task decomposition.} 
 Our contributions, grounded in the field of cognitive robotics, are : 
  a new  representation of complex actions enabling the exploitation of task decomposition and the proposition for teachers of 
 supplying information on the task hierarchy to learn compound tasks.

\section{Socially Guided Intrinsic Motivation}

We have presented two algorithms for active imitation learning based on intrinsic motivation. While  SGIM-ACTS learns several parametrised tasks by choosing its teachers and the timing of its requests, and SGIM-PB learns uses transfer of knowledge between tasks by learning the relationship between the parametrised tasks with the supplementary strategies of task decomposition by autonomous exploration and task decomposition imitation.  The main differences between SGIM-D, SGIM-ACTS and SGIM-PB are outlined and they are contrasted with autonomous learning algorithm IM-PB in Table \ref{tab:DiffAlgoSGIM}. 

\begin{sidewaystable}

\caption{Differences between SGIM-D, SGIM-ACTS, SGIM-PB and IM-PB in the learning strategies used, the imitation content and timing, and the modalities of transfer of knowledge.}
\label{tab:DiffAlgoSGIM}
\begin{tabular}{|p{2.3cm} |p{2.2cm} |p{5cm} |p{2.5cm} |p{2cm} |p{2cm}|p{2cm}|}
\toprule
\textbf{Algorithm} & \textbf{Action Representation} & \textbf{Strategies} \boldmath{$\sigma$} & \textbf{What to imitate} & \textbf{When to imitate} & \textbf{Who to imitate} & \textbf{Transfer of Knowledge} \\
\hline
IM-PB~\cite{Duminy2018PIICRC} &  parametrised continuous actions, task decomposition &  Autonomous action exploration, Autonomous outcome space exploration, autonomous task decomposition exploration & NA & NA & None & Cross-task transfer \\
\hline
SGIM-D~\cite{Nguyen2014AR} &  parametrised continuous actions & autonomous action space exploration, mimicry of an action teacher, emulation of an outcome teacher  & Teacher demo. of actions and outcomes & Predefined frequency & 1 teacher with correspondence problems & Mimicry, Emulation
\\
\hline
SGIM-ACTS~\cite{Nguyen2012PJBR} &  parametrised continuous actions & autonomous action space exploration, mimicry of action teachers, emulation of outcome teachers & Teacher demo. of actions and outcomes & Active request by the learner to the teacher & Several (imperfect) teachers & 
Mimicry, Emulation \\
\hline
SGIM-PB~\cite{Duminy2021AS} &  parametrised continuous actions, task decomposition & autonomous action space exploration, autonomous procedural space exploration, mimicry of an action teacher, mimicry of a task decomposition teacher  & Teacher demo. of actions and task decomposition & Active request by the learner to the teacher &  Several teachers & Cross-task transfer, Mimicry, Emulation \\
\bottomrule
\end{tabular}
\end{sidewaystable}

The various implementations and  experimental setups show that the robot benefits more from exploring directly the policy space for simple tasks, and by trying to decompose into subtasks for  sequential tasks. With our intrinsic motivation criteria, it can choose at each time step the appropriate exploration strategy, both in autonomous exploration strategies and in imitation learning strategies, and thus automatically composes its learning curriculum.

\begin{tcolorbox}
To summarise, I showed  that \textbf{the same formulation of the intrinsic motivation  is valid both for autonomous exploration and  for social guidance, may the demonstrations requested to teachers be low-level policies, goals or decomposition into subgoals, in multi-task learning setting with parametrised goals or hierarchical goals}. Thus, by considering that each choice of teacher and type of demonstration is a learning strategy, the intrinsic motivation for a learning strategy $\sigma$ and task $\omega$ can be modelled as 

 \begin{equation}
 im(\sigma, \omega) = \kappa(\sigma) * progress(\sigma, \omega) 
 \end{equation}
 
 where $progress$ is the empirical progress measured through the last episodes with strategy $\sigma$ and goal $\omega$, and $\kappa(\sigma)$ is the cost of the strategy, representing the availability of teachers, their willingness to interact with the robot ... 
\end{tcolorbox}
\newpage
\chapter{Application to Socially Assistive Robotics}
\label{sec:sar}

\section{A Robot Coach Proposing an Automatic Curriculum}

The algorithms for multi-task learning SGIM, CHIME and GARA automatically design their own learning curriculum, by deciding at each step which task to learn and the appropriate learning strategy. Such a learning curriculum can be used by intelligent tutoring systems for a robot coach to propose exercises personalised to each user. In \citep{Annabi20232IICDLI} we study an \textbf{intelligent tutoring system based on intrinsic motivation that chooses exercises to maximise the progress for each student}. The efficiency of this intelligent tutoring system relies on the knowledge about the relationship between the exercises, and the relationship between the skills. As with our study on hierarchically sequential tasks, some exercises train high-level skills (for instance multiplication in maths) that require other simpler skills to be mastered before (for instance addition in maths is a prerequisite before learning multiplication). We propose an algorithm to \textbf{uncover these pre-requisite relationships using causality analysis, based only data of student scores}. We therefore obtain a hierarchical relationship between exercises, but this time with respect to pre-requisite requirements.

These intelligent tutoring systems can be applied to socially assistive robots to personalise training programs for robot coaches.  

\section{Robot Coach for Physical Rehabilitation}

\subsection{Humanoid Socially Assistive Robot in a Clinical Trial}

As part of the Keraal experiment, funded by  EU FP-7 ECHORD++, for which I was the scientific coordinator, we proposed \textbf{a humanoid robot coach  for physical rehabilitation}.  \hl{The consortium includes Olivier Remy Neris, director of the rehabilitation department of CHRU Brest; the company Generation Robots; Mathieu Simonnet, specilaist in cognitive ergonomy at LEGO of IMT Atlantique, and Myriam Le Goff-Pronost, health economist at LaTIM, IMT Atlantique} .  Its goal is to entice motivation in patients while the patients exercises by themselves. This humanoid robot can entice motivation by its embodied presence and by the feedback it can give. In \citep{Devanne2017ICHRH}, we applied Gaussian Mixture Model (GMM) on Riemannian manifolds \citep{Zeestraten2017IRAL} for giving \textbf{feedback after analysing patients' movements} for three low-back pain exercises using a single RGB-D camera. In later works, we tested LSTM auto-encoder \citep{Nguyen2024IJCNN} and Spatio-Temporal Graph Convolutional Networks \citep{Marusic2023ECMR} to analyse the whole body movements. 

 In order to design our system and our HRI, we first choose an anthropomorphic robot platform, and led a psychological analysis of the target population in a co-design process \citep{Devanne2018IICRC}, \hl{in collaboration with Gilles Kermarrec, psychologist at  European Research Center for Virtual Reality and Research Center for Education, Learning and Didactics, European University of Brittany}.  Then, this system has been tested with 21 subjects, including 12 patients  in \textbf{clinical trials}  in a longitudinal study over daily sessions over three weeks for each patient. The system is assessed with medical criteria that there is non-inferiority of this system compared to classical care with therapists \citep{Blanchard2022BRI}.

\begin{figure}[tb]
\centering
\includegraphics[width= 0.6\textwidth]{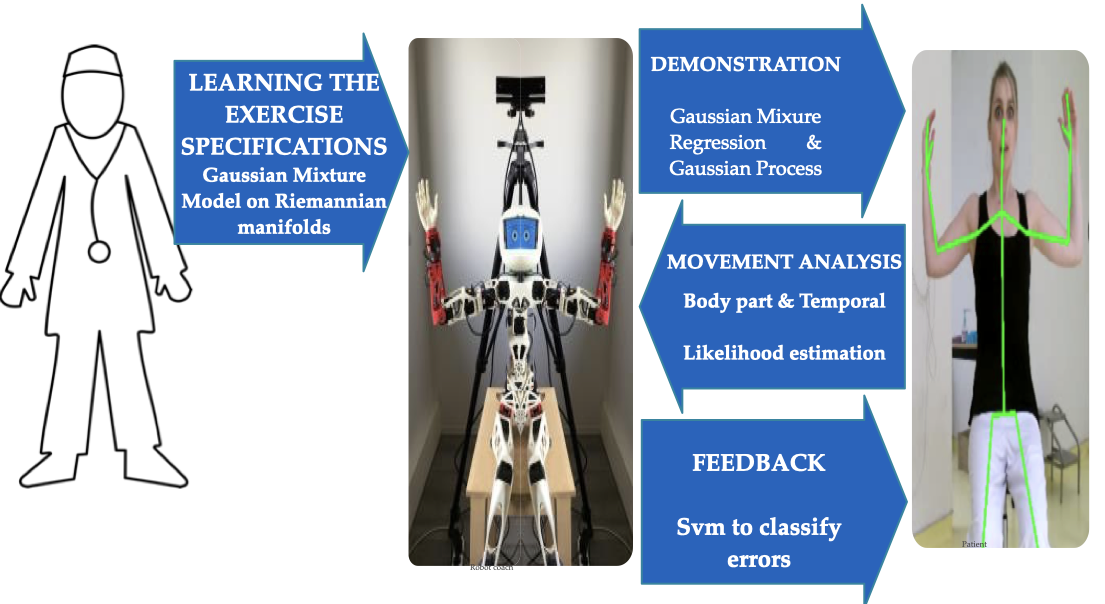}
\caption{ The robot learns the probabilistic model of exercises from demonstrations using GMM on Riemannian manifolds. The robot assesses the patient's movements   per body part and temporal segment by estimating the likelihood, and gives a feedback after classification of the error.}
\label{fig:Keraal}
\end{figure}

\subsection{Four Challenges for a Dataset}

This clinical trial enables us to identity four challenges for robot coaches and publish a medical low-back pain physical rehabilitation dataset for human body movement analysis  \citep{Nguyen2024IJCNN}. With the aim of rehabilitation using human body movement analysis, intelligent tutoring systems (ITS) need to analyse complex full-body exercises that can involve several parts of the body but not necessarily all parts of the body. The assessment algorithm should be able to understand which parts are important, and what are the ranges of freedom that are acceptable. The ITS should  encapsulate the tolerated variance for each joint and time frame. Thus, we have identified 4 challenges in rehabilitation movements analysis:

 \begin{enumerate}
 
     \item Rehabilitation motion assessment. The goal is to assess an observed motion sequence by detecting if the rehabilitation exercise is correctly performed or not. Thus each recording is labeled with a global evaluation: correct or incorrect, and also if the error is significant, medium or small.
    
     \item Rehabilitation error recognition. The goal is to classify the observed error among a set of known errors, so as to explain and give feedback.  Each recording has the label of the error and also if the error is significant, medium or small.

     \item Spatial localization of the error. In addition to recognizing the error, the goal is also to identify which body part is responsible of the error to draw the patient's attention to it. Each annotation indicates the body part causing the error.

     \item Temporal localization of the error. The goal is to detect the temporal segment where the detected error occurred along the sequence. Each annotation indicates the time window where the error occurs.
 \end{enumerate}

 While most rehabilitation datasets \citep{Leightley20152ASIPAASCA,Ar2014ITNSRE,Vakanski2018D,Dolatabadi2017P1ICPCTH,Miron2021D,Aung2016ITAC,Capecci2019ITNSRE} only provide annotations for the rehabilitation motion assessment, we propose a medical dataset recruiting 12 clinical patients carrying out low back-pain rehabilitation exercises over 3 weeks of rehabilitation program. The Keraal dataset  is composed of 2622 recordings , including 1881 recordings from patients. Each recording is composed of 3D Kinect skeleton positions and orientations, RGB videos, 2D skeleton data, and medical annotations by two medical annotators to assess the correctness, label and timing errors of each movement part.
 
 \subsection{Personalisation of the coaching to different morphologies}

\begin{figure}[tb]
\centering
\vspace{-1em}
\includegraphics[width= 0.6\textwidth]{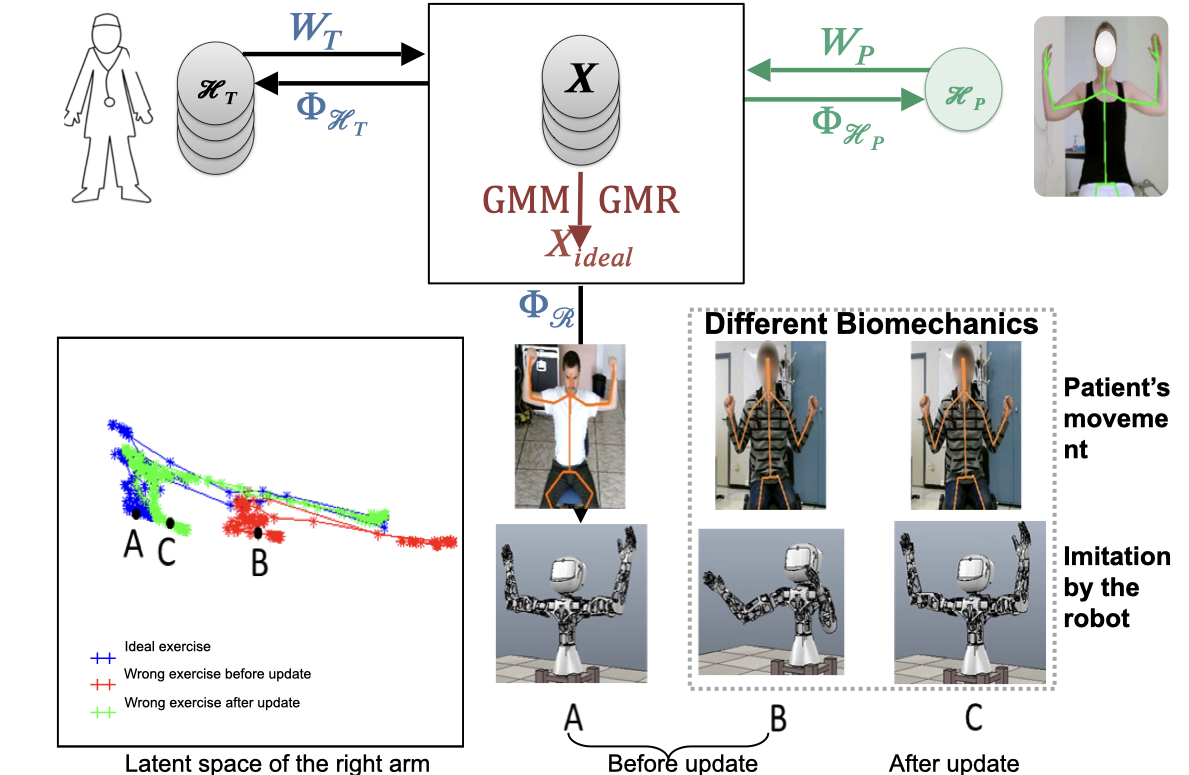}
\vspace{-0.5em}
\caption{The shared GP-LVM algorithms learn a latent space X from paired data from teacher's and a patient's movements, and the parameters $W$ and $\Phi$ to translate between each person's movement space and the latent space.  Using Gaussian Mixture Model (GMM) and Gaussian Mixture Regression (GMR), we translate the movement of a new patient to make the robot imitate.  Bottom right: illustration for three movements : the movement of a new patient, the corresponding movement by the robot. Bottom left: representation in the latent space of the right arm of the three movements: movement A is the ideal movement of the exercise by a patient with the same biomechanics as the teacher; movement B is the movement by a patient with a different biomechanics, before the model update.; movement C is the same as B  but after the model update.\\
Figure from \citep{Devanne2019CVE2W}.
}
\label{fig:SharedGPLVM}
\vspace{-0.6cm}
\end{figure}

Our current work seeks to personalise this coaching, by taking into account the physical limitations of each person. We proposed to use shared latent variables to translate the movement representation between the robot and the human motions from patients of different morphologies, representing patients who have different biomechanical constraints because of their pathologies or physical differences. In \citep{Devanne2019CVE2W}, the common representation was obtained using a shared Gaussian Process Latent Variable Model (shared GP-LVM).  Simulating mechanical contraints of joint angles, we could show that the robot could extend the model to these contraints and adapt the assessment of rehabilitation exercises to each patient's physical limitation,  by updating the inverse mapping matrix $W$, as shown in Fig. \ref{fig:SharedGPLVM}.

More recently, we started investigating how we can learn this common latent space when we do not have recordings of all exercises for the all biomechanical models. Indeed, in the absence of paired data, supervised data is not possible. In \cite{Annabi2024C2AICHI}, we explored unsuccessfully deep learning methods for unpaired domain-to-domain translation, that we adapt in order to perform human-robot imitation.

 The next steps would be to propose an intelligent tutoring system to personalise the training curriculum to each patient. We plan  toanalyse the exercises proposed by therapists to uncover their prerequisite relationships and to personalise the training curriculum, by applying the algorithms to uncover the pre-requisite relationship between exercises proposed in \citep{Annabi20232IICDLI}.  

\section{Imitation Game for ASD children}

In collaboration with the \hl{psychiatrist Nathalie Collot-Lavenne from} of CHRU Brest, we adapted a program of their daily care for \textbf{autism spectrum disorder (ASD) children to build a robot coach for imitation games}, building on the same movement analysis algorithms.

Indeed, while autism is an neurodevelopmental condition affecting social interactions and motor capacities, it has been shown that robots and ICT technologies can induce social and cognitive stimultation of children with ASD. This seems to be partly because they are more comfortable with explicit and predictive behaviors. Several studies have shown that imitating ASD children can be efficient in enhancing their social behaviour \citep{Nadel2005Imitationetaustisme}. These studies have been carried with caregivers as imitators of children's movements \citep{Sanefuji2009IMHJOPWAIMH,Katagiri2010RASD}. While most imitation studies have focused on facial expressions or sounds \citep{Dautenhahn2009ABB,Kozima2009IJSR,Pioggia2007R21IISRHIC,Jouen2017CAPMH}, few studies have focused on the imitation of movements. Development of motor control requires forming an internal model of action relying on the coupling between action (motor commands) and perception (sensory feedback). Critical to the development of social, communicative, and motor coordination behaviors, internal model of action accurately predicts the sensory consequences of motor commands \citep{Krakauer2007CNTIIBF}. This is why, we propose to examine how ASD children can learn to improve their motor control by a serious game. We propose to use a robot platform as they have been shown to arise their interest.  We therefore implemented a robot for an imitation game based on a nursery rhyme. Through a user experiment, we tested the application of the Gaussian Mixture Model to analyse  the movements of ASD children in the setting. The experiments were described in \citep{Nguyen2019WCTIWASD,Vallee2020IIWSRND}. These studies support the use of a humanoid robot as a theurapeutic tool to improve imitation skills in ASD children.

\chapter{The motivation to interact actively with tutors}

\begin{figure}[tb]
\centering
\vspace{-1em}
\includegraphics[width= \textwidth]{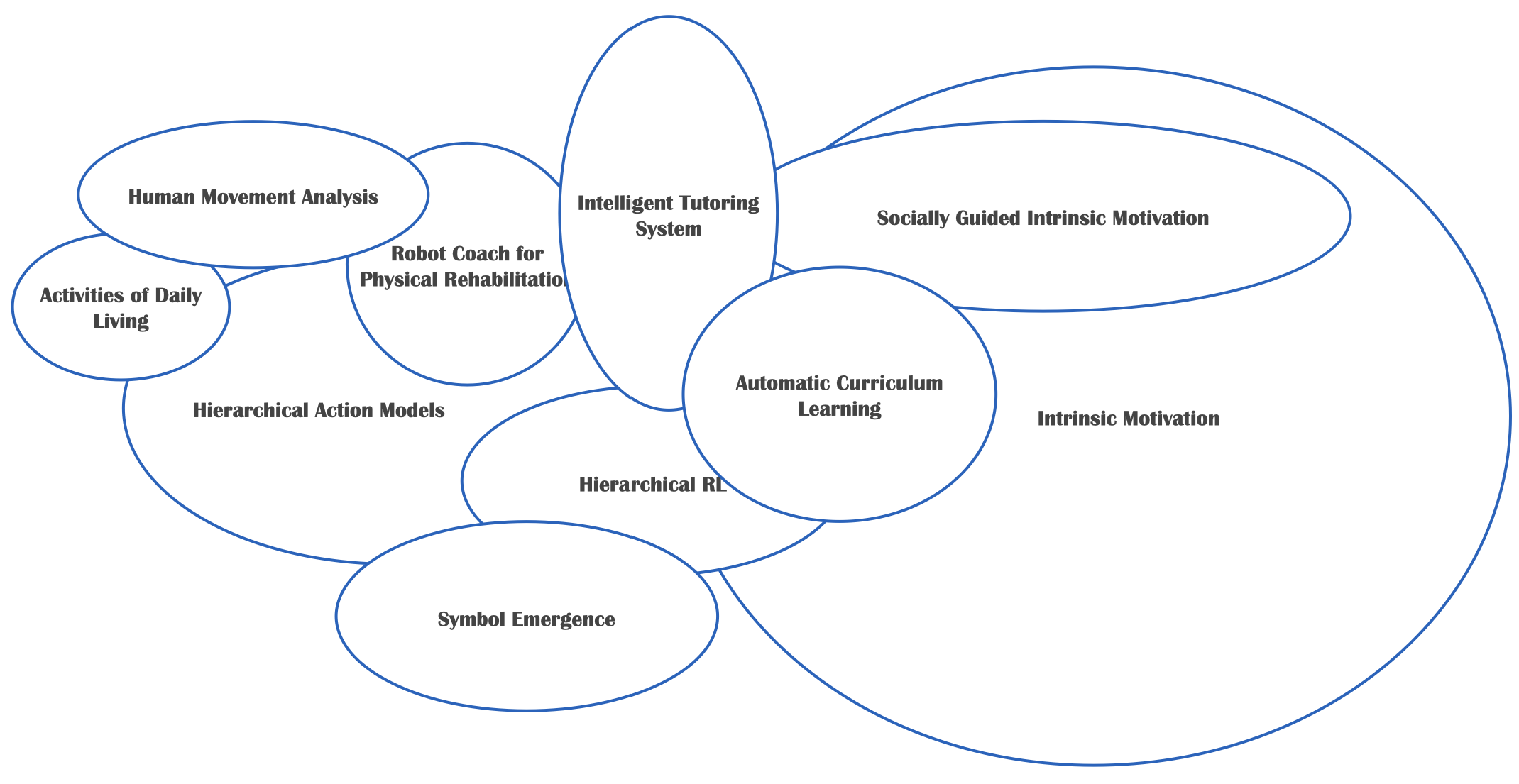}
\vspace{-0.5em}
\caption{Fields of our research projects
}
\label{fig:domains}
\vspace{-0.6cm}
\end{figure}

Our work lies at the intersection of several fields, which can be related as pictured in Fig. \ref{fig:domains}. We propose hierarchical action models through the study of activities of daily living and human movement analysis. The latter is applied for socially assistive robotics with a robot coach for physical rehabilitation. 

In parallel, hierarchical action models are also harnessed in reinforcement learning with algorithms of hierarchical reinforcement learning algorithms that rely on symbol emergence and the heuristic of intrinsic motivation to structure its learning strategy and derive automatically its learning curriculum. Most of all, we frame the field of socially guided intrinsic motivation, that extends the theory of intrinsic motivation to situations where the agent chooses aspects of its interaction with teachers. We formulate a reward function for choosing when, what, whom to imitate for multi-task learning, hierarchical learning and automatic curriculum learning.  
We aim to apply this theory for intelligent tutoring systems.

In Chapter 3, through the SGIM framework \citep{Nguyen2012PJBR,Duminy2021AS}, with my previous advisors and my collaborators, I proposed an \textbf{active imitation learning algorithm for the robot to actively decide  when, what and from whom to imitate, and when to learn autonomously, based on intrinsic motivation} \citep{Oudeyer2009FN,Schmidhuber2010ITAMD}.  
Nevertheless, the tasks solved remain quite simple, and have the disadvantage of relying only on low-level demonstrations by kinaesthetic  of movements or outcomes, instead of more natural means of communication with humans. In the future, we aim for more complex tasks such as tool use, and explore how this active imitation learning can use natural interactions. 

\section{Research problem}

Developmental psychology works such as \citep{Fagard2016FP} have started to study the influence of social guidance beyond the learning of simple motor skills, but have researched the learning of hierarchical skills such as tool use.  They highlight the existence of a zone of proximal development of children for sequential tasks. Even when each subtask is known, composing the subtasks can be a challenge. In \citep{Fagard2016FP}, children from 18 months of age succeeded in tool use thanks to demonstrations, while younger children could not benefit from it. Therefore, there is a pre-requisite to social learning. We argue that this pre-requisite is linked to an emergent representation of complex tasks that needs to be known by learning agent to enable it to map the observations to its representational structure.

\begin{tcolorbox}
The purpose of my future research project is to model both how  artificial agents can interact efficiently with tutors to learn in open-ended settings, and to model the motivations of humans to interact with tutors in a hierarchical learning setting. We investigate representations of sequential tasks and their emergence, as well as the refinement and the generalisation of my formulation of the motivational drive of learning agents to interact with tutors.
\end{tcolorbox}

We propose to investigate following two axes.

The first axis examines how a representation of complex tasks can enable the learning agent to understand the guidance of tutors. We hypothesise that an emerging discrete representation of our continuous environment and tasks is key to an efficient communication between the learner and the tutor.

The second axis examines the different factors of the motivational drive of learning agents to interact with tutors by analysing human interactions with tutors and by expanding the model of intrinsic motivation for learning to more drives and factors, integrating human-robot interaction methods, affective computing theories. For instance, we will expand beyond the intrinsic motivation based on learning progress and consider other intrinsic motivations such as homeostatic drives. Besides, we investigate other functions of imitation than learning : imitation can and is often mostly considered as a communication method \citep{Nadel2004IS}. The structures underlying our ability to imitate enable the emergence of both learning and communication abilities. For instance a tutor imitating the learner provides salient signals for his learning process, in which case imitation is both a means of a communication and of teaching \citep{Andry2001ITSCPSH}.

\section{Representations of tasks to enable social learning}

\subsection{Challenges}

To enable natural communication between robots and humans, we look for a \textit{learning process for robots to acquire a discrete representation of tasks in a continuous environment}, and \textit{how this discrete representation can be used by the robot to learn more efficiently with human guidance}. 
Indeed, symbol emergence is key for open-ended learning to tackle the curse of dimensionality and scale up to open-ended high-dimensional sensorimotor space, by allowing symbolic reasoning, compositionality, hierarchical organisation of the knowledge, etc. While symbol emergence has been recently investigated for the sensor data such as images and video, action symbolization can lead to a repertoire of various movement patterns by bottom-up processes, which can be used by top-down processes such as composition to form an action sequence, planning and reasoning for more efficient learning, as reviewed in \citep{Taniguchi2018ITCDS}, but also social guidance.

We will propose a learning algorithm where the robot learns multiple compositional tasks by deciding itself its learning strategy between reinforcement learning and social guidance and how to efficiently interact with tutors, by means of a symbolic representation of the tasks, emerging from its sensorimotor data. 
This line of research contributes to human-centered Artificial Intelligence and robotic learning.


\subsection{Learning a symbolic representation}
Human users can give teaching signals to robots by low-level kinaesthetic demonstrations of movements \citep{Billard2004RAS,Schaal2003PTRSLSBS}, but can also take a more natural means of communication on a higher level. High-level communication, such as language relies on symbols or a discrete abstract representation of space. How can a robot understand these words or symbols used by human communication ? How can a robot acquire such a discrete abstract representation of the sensorimotor data of the continuous environment ?

Besides, from the point of view open-ended learning, learning algorithms benefit immensely from the use of symbolic methods for goal representation as they offer ways to structure knowledge for efficient and transferable learning. 
However, the most existing Hierarchical Reinforcement Learning approaches relying on symbolic reasoning are often limited as they require a manual goal representation.
The challenge in autonomously discovering a symbolic goal representation is that it must preserve critical information, such as the environment dynamics. 
Recent works in hierarchical reinforcement learning \citep{Vezhnevets2017C, Nachum2019ICLRI2O2, Zhang2020, Li2022ICLR} have shown that 
learning an abstract goal representation is key to proposing semantically meaningful subgoals and to solving more complex tasks. In particular, representations that capture environment dynamics over an abstract temporal scale have been shown to provide interesting properties with regards to bounding the suboptimality of learned policies under abstract goal spaces \citep{Nachum2019ICLRI2O2,Abel2020PTTICAIS}, as well as efficiently handling continuous control problems. 
However, temporal abstractions that capture aspects of the environment dynamics \citep{Ghosh2019ICLR,Savinov2019ICLR,Eysenbach2019N,Zhang2020,Li2022ICLR} still cannot scale to environments where the pairwise state reachability relation is complex. This situation typically occurs when temporally abstract relations take into account more variables in the state space. The main limitation of these approaches is the lack of a spatial abstraction to generalise such relations over states.

Alternatively, other works \citep{Kulkarni2016ANIPS,Illanes2020PICAPS, Garnelo2016}) have studied various forms of spatial abstractions for goal spaces. These abstractions effectively group states with similar roles in sets to construct a discrete goal space. The advantage of such representation is a smaller size exploration space that expresses large and long-horizon tasks. 
In contrast to these algorithms that require varying levels of prior knowledge, our proposed algorithm STAR \citep{Zadem2024TICLR} gradually learns such spatial abstractions by considering reachability relations between sets of states.  While these representations have been obtained in simple maze environments, we will scale these algorithms to robotic setups in higher dimensions and with stochasticity.

\section{Different factors of the motivational drive of learning agents to interact with tutors}

 In the Socially Guided Intrinsic Motivation (SGIM) framework, I proposed a  common formulation of the intrinsic motivation valid both any sampling strategy, may it be autonomous exploration or  for social guidance, and may the demonstrations requested to tutors be low-level policies, goals or decomposition into subgoals. Thus, the intrinsic motivation for a learning strategy $\sigma$ and task $\omega$ can be modelled as  
 \begin{equation}
 im(\sigma, \omega) = \kappa(\sigma) * progress(\sigma, \omega) 
 \end{equation}
  where $progress$ is the empirical progress measured through the last episodes with strategy $\sigma$ and goal $\omega$, and $\kappa(\sigma)$ is the cost of the strategy. In the previous studies, this is was considered a constant set arbitrarily. However,  this cost representing the availability of teachers, their willingness to interact with the robot ...  but also other motivational drives of learners to interact with tutors.

To identify the motivational drives of learners to interact with tutors, I will analyse human behaviour in coaching situations in socially assistive robotics (SAR), by taking the use case of a robot coach for physical exercise. This work will take advantage of the framework laid in project Keraal, capitalising on the medical dataset collected during project Keraal. This analysis will enrich the formulation of the cost of each strategy, and will enable a more precise and automatic formulation of the cost coefficient as a function of the other motivational drives. It will enable us to generalise the formulation proposed, and sketch a common theory of the motivations to interact with tutors.

\section{Conclusion}
To tackle the problem of open-ended learning, my work in developmental robotics investigates the potential of social guidance to bootstrap the learning. My proposal provides a theoretical model enabling artificial agents to interact with tutors, by first investigating a prerequisite for efficient guidance : an appropriate representation of sequential tasks, and by second precising the formulation of the motivation criteria based on human-robot interaction studies with the use case of an intelligent tutoring system for a robot coach.  This research project is instrumental for study the notion of affordance. compositionality of actions, the co-development of tool-use and language, and the grounding of language. It will entail an alignment of AI to human representations.


\clearpage

\chapter{Selected Articles}

\begin{itemize}
\item \bibentry{Zadem2024TICLR}
\item \bibentry{Nguyen2024IJCNN}
\item \bibentry{Bouchabou2023MLITSWEP}
\item \bibentry{Duminy2021AS}
\item \bibentry{Nguyen2012PJBR}
\end{itemize}

\pagebreak

\chapter{Publication list of Sao Mai Nguyen}

Download available on \url{https://cv.hal.science/sao-mai-nguyen} and \url{http://nguyensmai.free.fr/Publications.html} \\

\begin{enumerate}
\section{Book chapter}
\item \bibentry{Thepaut2018RSSS} 

\section{Thesis }
\item \bibentry{Nguyen2013} 
\item \bibentry{Nguyen2010}

\section{Editorials}
\item \bibentry{Navarro-Guerrero2021ITCDS}

  
\section{Peer Reviewed Journals}
\item \bibentry{Bouchabou2023S}  {\small  \textit{(Impact Factor = 3.9, top 25 \% en Information Systems)}}
\item \bibentry{Bouchabou2023RODA}
\item \bibentry{Blanchard2022BRI}
\item \bibentry{Bouchabou2021E}   {\small  \textit{(Impact Factor = 2.9)}}  
\item \bibentry{Bouchabou2021S}  {\small  \textit{(Impact Factor = 3.9, top 25 \% en Information Systems)}} 

\item \bibentry{Duminy2021AS} {\small  \textit{(Impact Factor = 2.474)}} 
\item \bibentry{Nguyen2021KI}  
\item \bibentry{Duminy2019FN} {\small  \textit{(Impact Factor = 2.6)}}  
\item \bibentry{Nguyen2017KI} {\small  \textit{(h5 index = 15)}}
\item \bibentry{Kenigsberg2017DIJSRP} {\small  \textit{(Impact Factor = 1.7)}}
\item \bibentry{Thepaut2017AMPRARFR} 
\item \bibentry{Moulin-Frier2014FP} {\small  \textit{(Impact Factor = 1.1,top 25 \% en psychologie)}} 
\item \bibentry{Nguyen2014AR} {\small  \textit{(Impact Factor = 3.7)}}
\item \bibentry{Ivaldi2013TAMD} {\small  \textit{(Impact Factor = 5, top 25\% en IA)}}
\item \bibentry{Nguyen2012PJBR} {\small  \textit{(Impact Factor = 2.5)}}

\section{Conferences with Proceedings}
\item \bibentry{Zadem2024TICLR} (rank A*)
\item \bibentry{Nguyen2024IJCNN} (rank A)
\item \bibentry{Annabi2024C2AICHI} (HRI is rank A*) 
\item \bibentry{Bouchabou2024E2} accepted (ECAI is rank A)
\item \bibentry{Marusic2024I} 
\item \bibentry{Stoyanova2024LT} 
\item \bibentry{Gan2024HSI}
\item \bibentry{Zadem20232IICDLI}
\item \bibentry{Annabi20232IICDLI}
\item \bibentry{Bouchabou2023MLITSWEP} (ECML is rank A)
\item \bibentry{Marusic2023ECMR}
\item \bibentry{Marusic2023C2AICHI}  (HRI is rank A*) 
\item \bibentry{Bouchabou2021DLHAR} (IJCAI is rank A*) 
\item \bibentry{Mitriakov20202IISSSRRS} 
\item \bibentry{Mitriakov20202IICFSF}
\item \bibentry{Devanne2019ICSC} 
\item \bibentry{Manoury2019PICHI} 
\item \bibentry{Manoury2019IIRC} 
\item \bibentry{Duminy2018PIICRC}  
\item \bibentry{Devanne2019CVE2W}  (ECCV is rank A*)
\item \bibentry{Devanne2018IICRC}  
\item \bibentry{Devanne2017ICHRH}  
\item \bibentry{Duminy2016I2JIICDLER} 
\item \bibentry{Nguyen2016R2IISHRIC} 
\item \bibentry{Duminy2016I2JIICDLER}
\item \bibentry{Nguyen2013RLDM} 
\item \bibentry{Nguyen2013IICDLE} :
\item \bibentry{Nguyen20122IISRHIC}
\item \bibentry{Nguyen2012PPCRDC}
\item \bibentry{Nguyen2012IICHR}
\item \bibentry{Nguyen2011IICDL} 
\item \bibentry{Nguyen2011IWALIFHT} (IJCAI is rank A*)
\item \bibentry{Nguyen2010P1ICER}

\section{International Conferences with Peer Review}
\item \bibentry{Ji2024IOLWN}
\item \bibentry{Ji2024N2WBML}
\item \bibentry{Ji2024N2WCLPMPF}
\item \bibentry{Zadem2023IMOLWN} (Neurips is rank A*)
\item \bibentry{Nguyen2024A} (ECAI is rank A)
\item \bibentry{Annabi2023IMOL}
\item \bibentry{Zadem2023IMOL}
\item \bibentry{Gan2023SSAC}  
\item \bibentry{Zadem2022IMOLI2} 
\item \bibentry{Bouchabou2022II}
\item \bibentry{Vallee2020IIWSRND} 
\item \bibentry{Nguyen2019WCTIWASD}
\item \bibentry{Mitriakov2019}
\item \bibentry{Duminy2018WCUSLI}
\item \bibentry{Nguyen2016IWCR}
\item \bibentry{Nguyen2013WHLRSS} (RSS is rank A)
\item \bibentry{Nguyen2013WALRRSS} (RSS is rank A)
\item \bibentry{Moulin-Frier2013WIM}
\item \bibentry{Nguyen2012DLERI2IIC}
\item \bibentry{Nguyen2012SSDRCB}
\item \bibentry{Nguyen2009NIWSSCN}
 
\end{enumerate}

\clearpage

\bibliographystyle{plainnat}
\bibliography{/Users/mai/Documents/Recherche/Bibliographie/ensta}


\end{document}